\RequirePackage{fix-cm}
\PassOptionsToPackage{normalem}{ulem}
\documentclass{aidata}
\usepackage{amsmath,amssymb,array,tabularx}
\usepackage{algorithm,algpseudocode}
\usepackage{needspace}

\pretocmd{\section}{\FloatBarrier\Needspace{5\baselineskip}}{}{}
\pretocmd{\subsection}{\FloatBarrier\Needspace{4\baselineskip}}{}{}
\graphicspath{{fig/}}
\DeclareTextFontCommand{\textbf}{\bfseries}

\newcommand{\setresult}[2]{\expandafter\def\csname result@#1\endcsname{#2}}
\newcommand{\metric}[1]{\csname result@#1\endcsname}
\setresult{types}{485}
\setresult{pool}{100,000}
\setresult{physical}{83,618}
\setresult{runtime}{99,586}
\setresult{both}{83,365}
\setresult{recovery-delta}{0.0052}
\setresult{recovery-ci}{[0.0000, 0.0117]}
\setresult{recovery-pairs}{28}
\setresult{filter-delta}{-0.0036}
\setresult{filter-ci}{[-0.0071, -0.0004]}
\setresult{filter-pairs}{300}

\newcommand{\paperfigure}[3]{\centering\includegraphics[width=\linewidth]{fig/#1.pdf}}
\newcommand{\conceptfigure}[1]{\centering\includegraphics[width=\linewidth]{fig/#1.pdf}}
\title{RFW-S: Everyday Deformable Objects\\with Independent Simulation Verification}
\author[1,2,*]{Jiarui Guo}
\author[2,3,*,\dagger]{Saining Zhang}
\author[2]{Jiaying Wan}
\author[1]{Jiawei Weng}
\author[2]{Junhao Chen}
\author[1,2]{Rui Zhang}
\author[1]{Yuanbo Lyu}
\author[2]{Zhiheng Lee}
\author[3]{Peishuo Li}
\author[1,2]{Jingrun Du}
\author[4]{Weiqing Xiao}
\author[1]{Weiyi Lin}
\author[1,\ddagger]{Zenan Meng}
\author[1,\ddagger]{HU WEI}
\author[1]{Lin Qu}
\author[2]{Zongzheng Zhang}
\author[2]{Hao Zhao}
\author[5,\ddagger]{Chaojian Li}

\affiliation[1]{Alibaba Group}
\affiliation[2]{Tsinghua University}
\affiliation[3]{Nanyang Technological University}
\affiliation[4]{Nanjing University}
\affiliation[5]{The Hong Kong University of Science and Technology}

\contribution[*]{Equal contribution}
\contribution[\dagger]{Project leader}
\contribution[\ddagger]{Corresponding authors.}
\checkdata[Email]{\{mengzenan.mzn, kongwang\}@alibaba-inc.com, chaojian@ust.hk}
\checkdata[Project Page]{\url{https://roboflywheel-com.github.io/RoboFlywheel-Soft/}}
\abstract{
Large-scale robotic simulation requires diverse deformable objects with executable physical properties and realistic interaction behaviors. However, existing 3D asset collections primarily emphasize visual diversity, while scalable physical construction and verification remain challenging. We introduce RFW-S, a large-scale dataset of 83,365 simulation-verified assets spanning 485 everyday categories across five structural families: soft solids, cloth sheets, shells, garments, and bags. We develop an automated pipeline integrating reference-guided 3D generation, structure-aware physical modeling, and LLM-assisted parameter estimation and calibration. To ensure physical reliability, we introduce an independent simulation verification framework that evaluates object behaviors under interactions distinct from those used for parameter calibration. Extensive experiments demonstrate the effectiveness of independent verification and characterize deformable responses under different physical configurations. We further evaluate robotic grasping in terms of both task success and object integrity, and demonstrate the physical realism of our simulated objects through sim-to-real replay of fragile-object manipulation. RFW-S provides a scalable foundation for physically grounded simulation and embodied robotic learning.
}
\begin{document}
\maketitle
\noindent\begin{minipage}{\linewidth}
\centering
\includegraphics[width=0.80\linewidth]{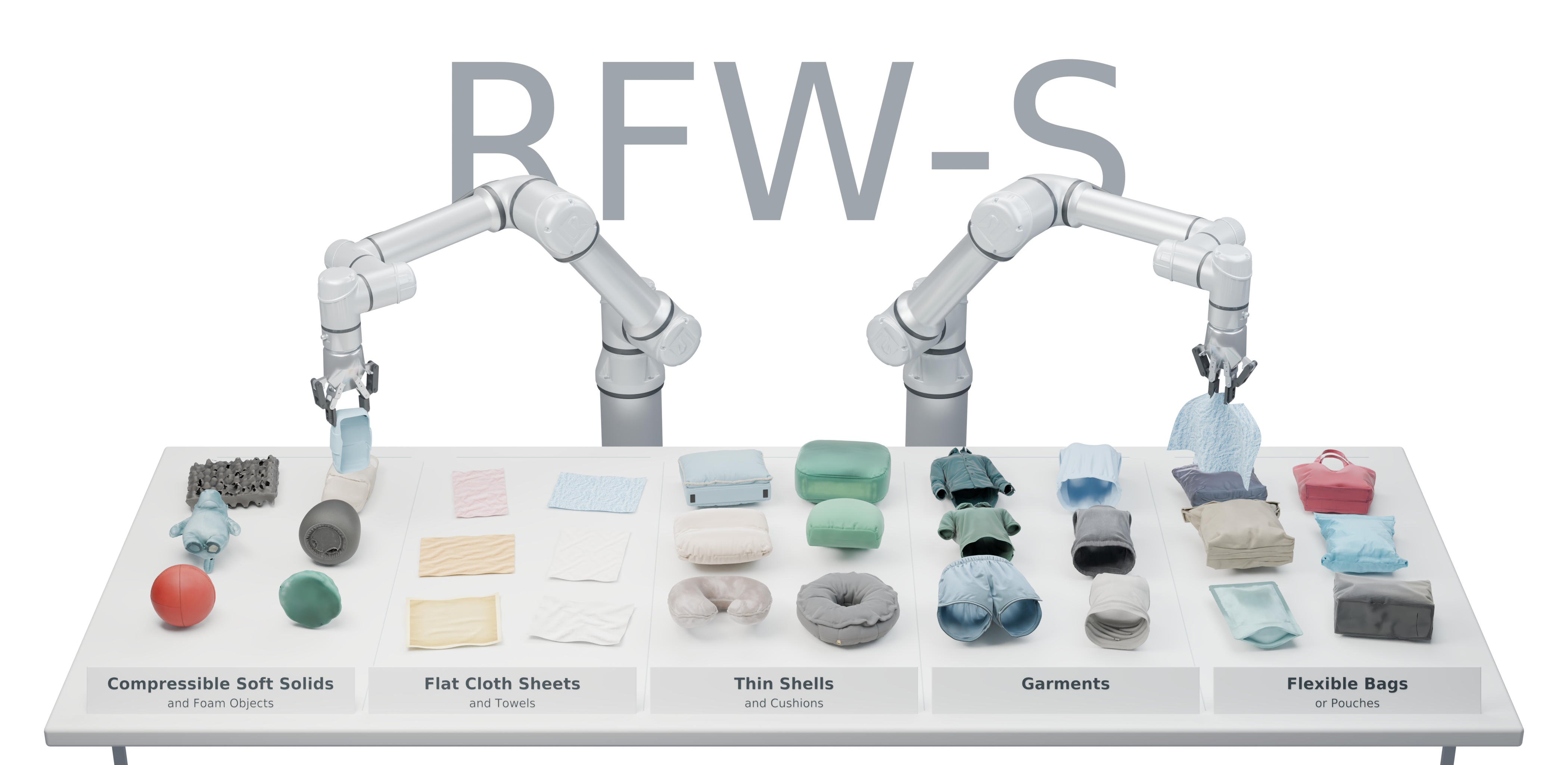}
\captionof{figure}{\label{fig:teaser}\textbf{Everyday deformable objects for robot interaction.}
RFW-S spans 485 categories across soft solids, cloth sheets, shells, garments,
and bags. This conceptual rendering illustrates the resource's scope.}
\end{minipage}\par
\section{Introduction}
\label{sec:introduction}
Robots operating in everyday environments must interact with objects whose
shape changes under contact: a sponge compresses during grasping, a towel
drapes over a surface, and a bag folds as it is lifted. These responses
affect how contact is established, whether a grasp succeeds, and whether an
object remains intact. Simulation offers a way to study such interactions
across object types and physical conditions, but doing so requires a diverse
collection of deformable assets with executable, tested physical behavior.

Existing resources provide important parts of this foundation. Large 3D
collections offer broad visual diversity \citep{deitke2023objaversexl}, while
clothing datasets and deformable simulation environments provide specialized
geometry and interaction settings
\citep{bertiche2020cloth3d,lin2020softgym}. Recent systems further connect
asset construction to physics: ManiTwin verifies rigid-object grasps,
PhysX-Omni generates assets across physical types, and Agentic Real2Sim
reconstructs recorded interaction episodes
\citep{wang2026manitwin,cao2026physxomni,chen2026agenticreal2sim}.
For everyday deformables, the remaining challenge is to combine broad
semantic coverage with physical configurations that can be reused and
assessed beyond the interactions used to construct them.

This challenge extends beyond assigning material parameters to a generated
mesh. Deformation depends jointly on geometry, structural constraints,
material controls, and contact. A filled soft object needs interior
material, an open bag needs a preserved cavity and opening, and a garment
needs a surface representation that permits folding. An unsuitable
representation can prevent the intended response even when its material
parameters are adjusted. Moreover, a configuration that runs without
numerical failure may still deform incorrectly, and one that satisfies a
calibration interaction may fail under a different stimulus. Scaling asset
generation, therefore, requires both structure-aware physical construction
and behavioral evaluation that is separate from candidate selection.

We introduce \textbf{RoboFlyWheel-Soft (RFW-S)}, a resource of
\textbf{83,365 simulation-verified assets spanning 485 everyday categories}
across five structural families: soft solids, cloth sheets, shells,
garments, and bags (Figure~\ref{fig:teaser}). We construct a category-balanced
pool of 100,000 assets through reference-guided generation, adapt their
geometry to compatible simulation representations, and use a multimodal
language model to estimate effective physical parameters. Simulation
feedback guides bounded corrections to supported controls. Each asset links its executed configuration to recorded behavioral evidence.

The central design is to separate calibration from final verification.
Once calibration selects a candidate, we freeze its geometry,
representation, parameters, and numerical settings, then evaluate it with
a distinct interaction. A separately scheduled
runtime and behavioral contract check evaluates execution, finite state,
and response criteria under a drop probe. In total, 83,365 assets pass both physical verification and the runtime and behavioral contract check, spanning all 485 everyday object categories.

We evaluate RFW-S through controlled physical tests, robotic grasping, and sim-to-real experiments. Our results show that material tuning alone cannot replace structural adaptation, and that calibration success does not guarantee valid behavior under a different interaction, motivating independent verification. Grasping experiments expose the gap between task completion and object integrity. Finally, replaying simulated gripper closure commands on a real robot reproduces both breakage and intact transfer, supporting the physical realism of our fragile-object simulation.

Our contributions are threefold:
\begin{itemize}
\setlength{\itemsep}{0pt}
\setlength{\parskip}{0pt}
\item \textbf{A large-scale dataset of everyday deformable objects.}
RFW-S provides 83,365 simulation-verified assets spanning 485 categories and
five structural families, with executable physical configurations and
associated behavioral evidence.
\item \textbf{A scalable pipeline for physically verified asset construction.}
We combine structure-aware modeling, language-model-assisted parameter
estimation, and simulation-guided calibration with independent interaction
tests that screen for failures beyond calibration.
\item \textbf{Physical and downstream validation.}
Controlled studies demonstrate the importance of structural adaptation and
independent verification. Robotic grasping evaluates task success alongside
object integrity, while sim-to-real replay reproduces both breakage and
intact transfer, supporting the realism of fragile-object simulation.
\end{itemize}

\section{Related Work}
\label{sec:related}
\paragraph{Resources and physical generation.}
Objaverse-XL provides broad visual coverage \citep{deitke2023objaversexl};
ABO, OmniObject3D, and Objaverse supply textured or annotated objects
\citep{collins2022abo,wu2023omniobject3d,deitke2023objaverse}.
CLOTH3D, ClothesNet, and GarmentCodeData organize garment geometry
\citep{bertiche2020cloth3d,zhou2023clothesnet,korosteleva2024garmentcodedata}.
SoftGym, PlasticineLab, DaXBench, and GarmentLab supply deformable interaction
environments \citep{lin2020softgym,huang2021plasticinelab,chen2023daxbench,lu2024garmentlab};
BEHAVIOR-1K connects household objects and activities \citep{li2024behavior1k}.
DeformableRavens supplies goal-conditioned tasks across cables, fabrics, and
bags \citep{seita2021deformableravens}. Simulation-oriented generation includes
Seed3D and PhysX-Anything \citep{feng2025seed3d,cao2026physxanything}.
Table~\ref{tab:positioning} compares resource units, coverage, physical
construction, and evaluation evidence. ManiTwin-100K provides 512 rigid-object
categories and simulation-verified grasps \citep{wang2026manitwin}.
PhysXVerse spans over 2.9K general categories, supporting physical generation
and behavior-based evaluation \citep{cao2026physxomni}.
RFW-S focuses its taxonomy on everyday deformables and reports category
retention after configuration-specific physical verification and RC.

\begin{table}[htbp]
\caption{\label{tab:positioning}\textbf{Resources and their physical evidence.}
Reported units and category definitions differ across resources. Panel B
describes the published construction and evaluation protocol. RFW-S tests a
frozen configuration beyond its calibration interaction. Citations are given
in the accompanying text.}
\centering\footnotesize\setlength{\tabcolsep}{3pt}
\begin{tabularx}{\linewidth}{@{}>{\raggedright\arraybackslash}p{.20\linewidth}>{\raggedright\arraybackslash}p{.22\linewidth}>{\raggedright\arraybackslash}p{.18\linewidth}>{\raggedright\arraybackslash}X@{}}
\toprule
\multicolumn{4}{@{}l}{\textbf{A. Scale, semantic organization, and physical scope}} \\
Resource & Reported scale / unit & Category organization & Object scope \\
\midrule
Objaverse-XL & 10M+ 3D objects & Diverse source collections & Broad visual geometry \\
CLOTH3D & 11.3K garments; 2.1M time samples & Six garment groups & Simulated clothing \\
SoftGym & 10 environments & Task-defined objects & Rope, cloth, fluid \\
ManiTwin-100K & 100K 3D objects & 512 categories & Rigid graspable objects \\
PhysXVerse & 8.7K+ 3D assets & 2.9K+ general categories & Rigid, deformable, articulated \\
Agentic Real2Sim & 100 rigid episodes evaluated; qualitative extensions & Episode-defined objects & Rigid, deformable, humanoid \\
\textbf{RFW-S} & \textbf{100K pool; 83,365 jointly verified} & \textbf{485 everyday categories} & \textbf{Five deformable families} \\
\bottomrule
\end{tabularx}
\par\smallskip
\begin{tabularx}{\linewidth}{@{}>{\raggedright\arraybackslash}p{.20\linewidth}>{\raggedright\arraybackslash}p{.22\linewidth}>{\raggedright\arraybackslash}p{.27\linewidth}>{\raggedright\arraybackslash}X@{}}
\toprule
\multicolumn{4}{@{}l}{\textbf{B. Parameters, validation protocol, and associated evidence}} \\
Resource & Physical construction & Reported evaluation & Evidence / reusable output \\
\midrule
Objaverse-XL & Source 3D geometry & Visual-generation evaluation & Deduplicated objects, renders \\
CLOTH3D & Blender fabric presets & Cloth motion on body sequences & Meshes, vertex motion, fabric labels \\
SoftGym & Parameterized FleX environments & Manipulation-policy rollouts & Environments, state and reward \\
ManiTwin-100K & Inferred size, mass, friction & Grasp stability and slide resistance & Mesh, properties, verified grasp poses \\
PhysXVerse / Omni & Part-wise physical attributes & Human annotation review; free-fall / water-drop scoring & Physical assets, six-attribute benchmark \\
Agentic Real2Sim & Inferred episodic simulation & Replay-based candidate judging & Executable twins, replay observations \\
\textbf{RFW-S} & \textbf{Representations and effective solver values} & \textbf{Calibration, independent probe, runtime and behavioral contract check} & \textbf{Geometry, parameters, reports, correction traces} \\
\bottomrule
\end{tabularx}

\end{table}

\paragraph{Physical inference and feedback.}
Cerberus jointly parses semantic categories, affordances, and attributes
\citep{cerberus2021}; PUGS predicts physical properties through a Gaussian
splatting representation \citep{pugs2025}.
Language priors and video observations support parameter inference
\citep{zhai2024nerf2physics,li2023pacnerf,jiang2025phystwin}.
Vid2Sid and Swim2Real use visual feedback for system identification
\citep{qiu2026vid2sid,qiu2026swim2real}; Agentic Real2Sim assembles executable
episodic twins \citep{chen2026agenticreal2sim}.
RFW-S uses calibration feedback during asset construction, then records a
separate final test. Its parameter comparisons use shared probes and evaluation budgets.

\paragraph{Deformable manipulation and damage.}
DefGraspSim evaluates deformable grasps; stress-guided learning and
OopsieVerse study fragile interaction and damage-aware outcomes
\citep{huang2022defgraspsim,ikemura2025gentle,balaji2026oopsieverse}.
TopoCut addresses damage-driven cutting \citep{wang2025topocut}.
Our grasping evaluation uses a deformation-gradient-based proxy to expose
the distinction between completing a task and preserving integrity.

\begin{figure}[!t]
\centering\begin{minipage}{0.78\linewidth}
\paperfigure{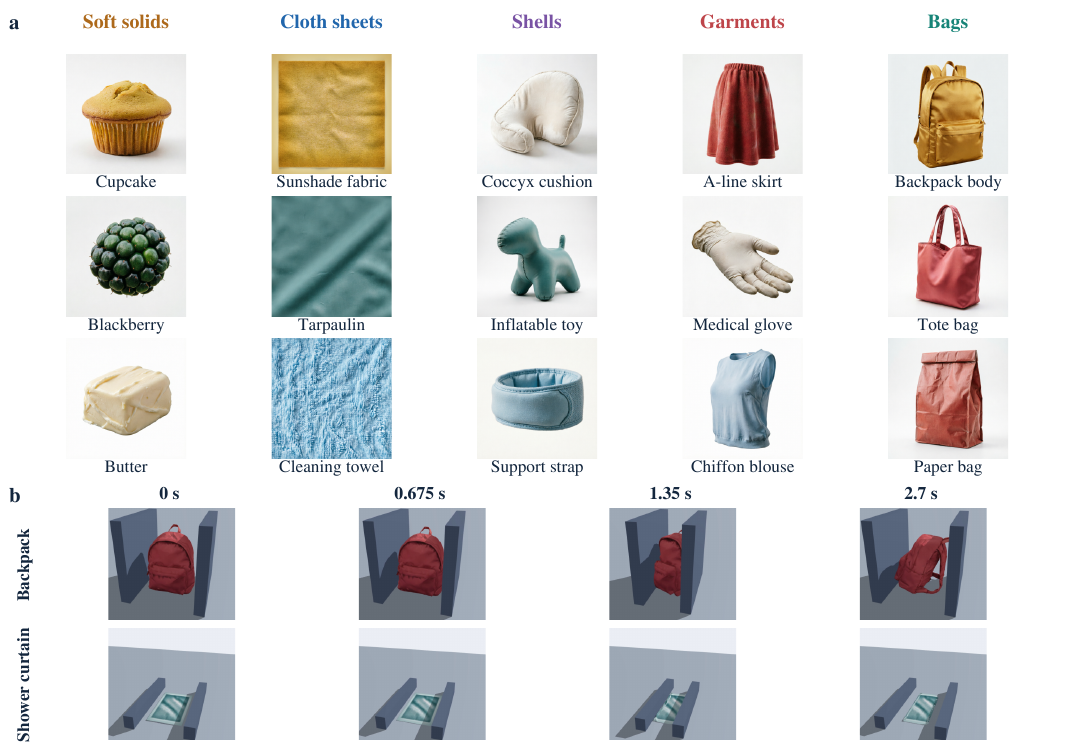}{4in}{Everyday assets and simulation sequences}
\end{minipage}
\caption{\label{fig:asset-examples}\textbf{Reference appearances and recorded simulation.}
Fifteen reference appearances sample the unfiltered asset pool across its
five structural families. Two recorded simulation sequences show compression
and release at matched physical times within each sequence. Reference images convey intended appearance, while
simulation frames show the executed representation.}
\end{figure}

\begin{table}[!b]
\caption{\label{tab:dataset}\textbf{Full-pool census.} Physical verification and
the runtime and behavioral contract check (RC) use the same 100,000 assets.
``Both'' is their configuration-matched
intersection; all categories remain represented.}
\centering\small\begin{tabular}{lrrrrr}
\toprule
Family & Types & Pool & Physical & RC & Both \\
\midrule
Soft solids & 168 & 34,641 & 21,112 & 34,502 & 21,103 \\
Cloth sheets & 89 & 18,347 & 18,308 & 18,097 & 18,064 \\
Shells & 63 & 12,990 & 11,415 & 12,990 & 11,415 \\
Garments & 94 & 19,386 & 19,055 & 19,386 & 19,055 \\
Bags & 71 & 14,636 & 13,728 & 14,611 & 13,728 \\
\textbf{Total} & \textbf{485} & \textbf{100,000} & \textbf{83,618} & \textbf{99,586} & \textbf{83,365} \\
\bottomrule
\end{tabular}

\end{table}

\begingroup
\let\FloatBarrier\relax
\section{The RFW-S Dataset}
\label{sec:dataset}
\subsection{Coverage and resource units}
\endgroup
The pool contains \metric{pool} assets in \metric{types} canonical everyday
categories. Figure~\ref{fig:asset-examples} illustrates their reference
appearances and recorded simulation responses. A category denotes an object type; a structural family groups
representation requirements. The five families contain 168 soft-solid,
89 cloth-sheet, 63 shell, 94 garment, and 71 bag categories
(Table~\ref{tab:dataset}). Appearance and parameter variations retain the
same category identity. Generation assigns 206 assets to 395 categories and
207 to the remaining 90, providing near-uniform research coverage. After both
checks, every category retains between 3 and 207 assets.

\Needspace{7\baselineskip}
Four units recur in the analysis. A \emph{category} is a semantic type, an
\emph{asset} is a generated object with a persistent identity, a
\emph{configuration} specifies its geometry, representation, parameters, and
solver settings, and an \emph{episode} executes a configured interaction.
Repeated episodes and parameter variants can share the same underlying
asset. Category counts describe semantic breadth; asset counts describe
the pool; episode counts describe the amount of evaluation. We preserve
these distinctions when reporting repeated method and budget comparisons.

\subsection{Asset contents and provenance}
\begin{figure}[htbp]
\paperfigure{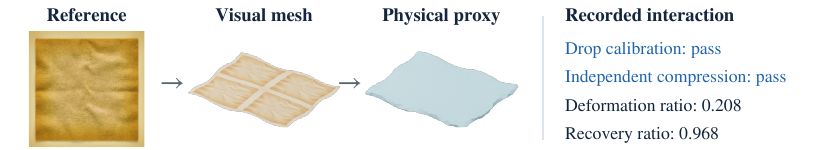}{1in}{An asset and its verification evidence}
\caption{\label{fig:asset-chain}\textbf{One asset from appearance to evidence.}
The reference, visual mesh, and physical-proxy render belong to the same
sunshade-fabric asset. The last panel reports that asset's calibration and
independent compression outcomes.}
\end{figure}

Each asset links semantic identity and reference appearance to visual
geometry, a simulation representation, scale, effective parameters, and
verification records. Figure~\ref{fig:asset-chain} shows a concrete example.
Visual and physical meshes may differ in topology; their association and
coordinate transforms are recorded. Numerical scale normalization is stored
separately from intended semantic size. The physical record identifies the
executed solver controls, including clamps and representation-dependent maps.

Reference appearances, generated visual meshes, and simulated proxies serve
different purposes. Reference images specify intended appearance; the visual
mesh provides the generated geometry; the physical representation supports
the executed solver. Figure~\ref{fig:asset-examples} shows reference images
and recorded interactions, while Figure~\ref{fig:asset-chain} links these
stages for a single asset.

\subsection{Verification scope and retained subsets}
\emph{Physical verification} evaluates calibration and independent behavioral
criteria for the selected configuration. A separately scheduled
\emph{runtime and behavioral contract check} (RC) requires both completed
execution with finite state and satisfaction of the behavior criteria under
its drop probe. In the archived records, 99,941 assets pass execution and
finiteness checks; 355 of these fail the behavior criteria, leaving 99,586
RC passes. The remaining 59 fail the numerical check.
The verified intersection matches asset and configuration identities:
\begin{equation}
\mathcal D_{\cap}=\{a\in\mathcal D_{\rm pool}:V_{\rm physical}(a)=1
\ \land\ V_{\rm RC}(a)=1\}.
\label{eq:intersection}
\end{equation}
The implementation uses Genesis~1.3.3 \citep{genesis2024}, with position-based
dynamics (PBD) and material point method (MPM) representations.
Verification denotes usability under these solver and interaction conditions.
The parameters are simulator-effective values, inferred during construction.

\begin{figure}[htbp]
\paperfigure{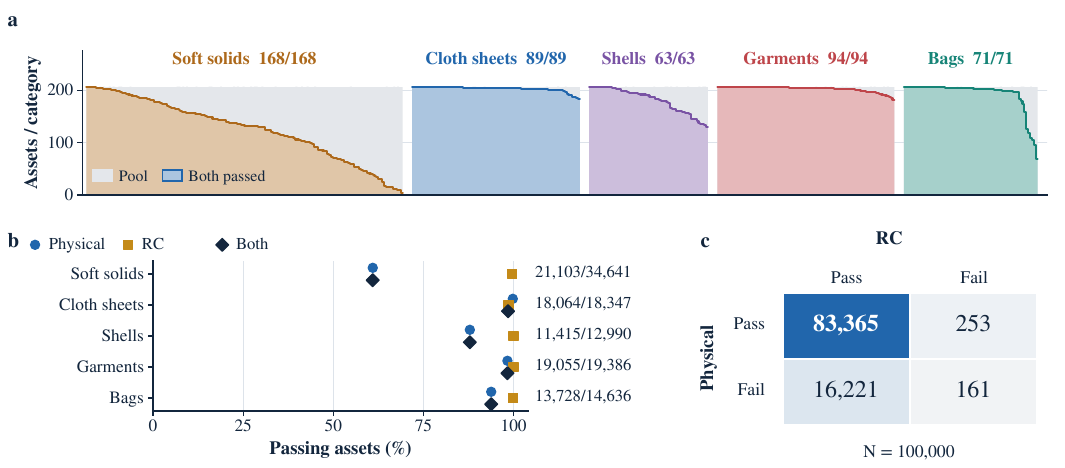}{2.2in}{Category coverage and verification}
\caption{\label{fig:coverage}\textbf{Coverage retained after verification.}
(a) Counts across all 485 categories. (b) Family pass proportions.
(c) Exact overlap: 83,365 pass both checks, 253 only physical verification,
16,221 only RC, and 161 neither. RC requires execution, finiteness, and
behavioral contract acceptance. Counts are a full census.}
\end{figure}

The two checks yield complementary screening outcomes. A total of 253
configurations pass physical verification but fail RC, while 16,221 pass RC
but fail physical verification, showing that passing either check alone is
insufficient for full acceptance. Their intersection yields 83,365 verified
assets spanning all 485 categories. Table~\ref{tab:dataset} summarizes asset
counts before and after filtering across structural families, while
Figure~\ref{fig:coverage} details retention at the category level.

\section{From Generation to Verified Simulation}
\label{sec:method}
Construction separates representation choice, effective parameterization,
candidate selection, and verification. Calibration observations can influence
the candidate that is selected; the independent probe evaluates that
selection after its geometry, parameters, and numerical settings are frozen.
The separately scheduled RC outcome is joined by configuration identity
when forming the usable subset.
\subsection{Coverage-aware generation and structural adaptation}
\begin{figure}[htbp]
\conceptfigure{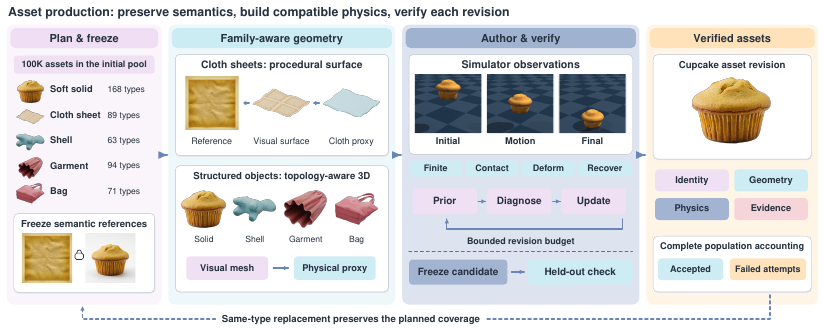}
\caption{\label{fig:pipeline}\textbf{Resource construction and final verification.}
Coverage-aware generation supplies geometry and reference appearances.
Structural adaptation and effective parameters make assets executable.
Calibration supports corrections; independent testing and the RC probe
provide evidence for the resulting configuration.}
\end{figure}

A canonical category and appearance specification condition Qwen-Image-2.0
reference generation \citep{zhao2026qwenimage2}. References guide TRELLIS.2
geometry generation \citep{xiang2025trellis2} and family-specific construction.
Thin sheets can be procedural; structured objects retain their generated
three-dimensional form. Geometry retries remain assigned to the same category
and selected reference, maintaining the intended allocation before filtering
(Figure~\ref{fig:pipeline}).

\label{sec:representation}
Structure determines the representation: filled objects require interior
material, bags need meaningful openings, and garments need surfaces that can
fold. We use filled-volume, sheet, cloth-garment, shell-garment, bag-shell,
reinforced-shell, and sealed-membrane profiles. Profiles determine compatible
solvers and constraints; a filled beanbag can use MPM despite its semantic
``bag'' label. Volume construction uses interior filling and surface
reconstruction. Surface adaptation removes small components and duplicated
vertices while preserving openings. Connectivity, finite vertices, boundary
edges, and repairs accompany the asset.

Semantic family guides this choice but does not uniquely determine a solver.
An open fabric bag and a filled beanbag can require different representations
despite sharing an everyday category family. The record therefore retains
the proposed route and the mode actually executed. Representation diagnosis
precedes interpretation of a parameter change: a control can only affect
behavior when the selected solver and adapter implement it.

Garment adaptation addresses a recurrent failure: strong rest-curvature
support can preserve a generated upright pose even with softer material.
The flexible cloth profile reduces this support while retaining stretch
resistance and self-collision. The matched configuration study in
Section~\ref{sec:structure-results} measures the resulting behavior.

\subsection{Effective parameters and calibration feedback}
\label{sec:feedback}
\begin{figure}[htbp]
\conceptfigure{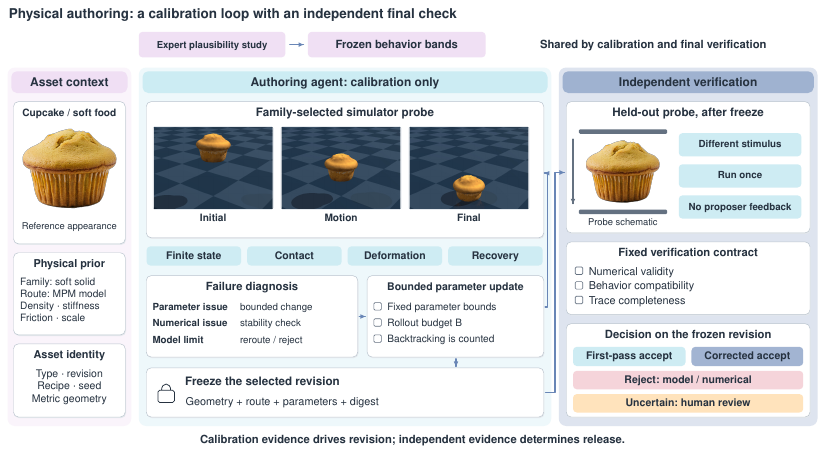}
\caption{\label{fig:agent}\textbf{Calibration feedback and independent testing.}
Diagnostics guide bounded updates through supported solver controls.
The selected geometry, route, and parameters are frozen before a distinct
probe. Expert judgments inform behavioral criteria; numerical validity
remains a separate requirement.}
\end{figure}

Given geometry $x$ and representation $r$, Qwen (API identifier
\texttt{qwen3.8-max}) estimates $\theta_0$ from reference images and semantic
cues. The adapter maps it to executable controls,
$\theta_{\rm eff}=A_r(\theta_0)$. A calibration probe
$S_r(x,\theta_{\rm eff};u_{\rm cal})$ measures contact, finite state,
deformation, recovery, and residual motion where applicable
(Figure~\ref{fig:agent}). Diagnostics distinguish parameter mismatch,
absent contact, numerical failure, and unsupported response. The same model
proposes bounded corrections from structured text summaries of calibration.
Iteration stops at acceptance
or the candidate budget; Appendix~\ref{app:backbones} gives the model settings.

Effective controls matter: a proposed viscosity field that is not passed to
the MPM material leaves the response unchanged. Local diagnostics instead
identify active controls, including yield stress for supported elastoplastic
representations. Observation duration also matters: a short window can
confound slow settling with persistent motion. The records therefore retain
the executed parameter map, time step, substeps, and observation duration
(Appendix~\ref{app:diagnostics}).

\subsection{Model roles and configuration}
\label{sec:model-roles}
Table~\ref{tab:model-config} gives the physical-construction model roles and
sampling settings. The Qwen stages use API inference without task-specific
fine-tuning. The recorded physical-prior model identifier and revision are
both the API alias \texttt{qwen3.8-max}. Local text-only controls use this
alias as a configurable default. Appendix~\ref{app:backbones} describes
the configuration and the model information retained in the records.

\begin{table}[htbp]
\caption{\label{tab:model-config}Model roles in physical construction and local
controls. Temperatures specify inference sampling.
The local controls' backend is specified by their transport configuration.}
\centering\small\setlength{\tabcolsep}{4pt}
\begin{tabularx}{\linewidth}{@{}p{0.23\linewidth}p{0.21\linewidth}Xr@{}}
\toprule
Component & Model / update & Input to proposer & Temp. \\
\midrule
Physical prior & \texttt{qwen3.8-max} & Reference images, semantics, parameter catalog & 0.1 \\
Production correction & \texttt{qwen3.8-max} & Prior, current candidate, calibration metrics and diagnosis & 0.2 \\
Local text-only controls & \texttt{qwen3.8-max} & Structured parameter JSON; no images or video & 0.2 \\
Numeric agent & Mechanical rules & Numerical calibration residuals & --- \\
\bottomrule
\end{tabularx}
\end{table}

\subsection{Frozen-candidate verification and behavioral criteria}
\label{sec:verification}
After calibration, geometry, representation, parameters, and numerical
settings are fixed. A distinct probe evaluates the candidate, for example
with lateral compression and release following drop calibration. This final
test records an acceptance or rejection of the frozen configuration:
\begin{equation}
V_{\rm physical}=G_{\rm execution}\land G_{\rm cal}\land G_{\rm independent}.
\label{eq:physical}
\end{equation}
Records distinguish first-candidate acceptance, acceptance after correction,
model-limit rejection, numerical rejection, and unresolved judgment.
A corrected acceptance requires an initial failure, a configuration change,
and an independent pass.

Algorithm~\ref{alg:verification} summarizes candidate selection and final
testing. Independent-test outcomes are recorded after calibration ends.
The RC probe is scheduled over the full pool and its result is joined by
configuration identity. Appendix~\ref{app:criteria} defines the response
metrics and acceptance bands.

\begin{algorithm}[htbp]
\caption{Candidate selection and independent verification}
\label{alg:verification}
\begin{algorithmic}[1]
\Require Asset geometry and context, calibration probe, independent probe, candidate budget
\State Construct a compatible representation and map the physical prior to effective controls
\While{calibration budget permits and the candidate is not accepted}
  \State Execute calibration; record the response, diagnosis, and complete configuration
  \If{calibration criteria are satisfied}
    \State Select the current candidate and stop calibration
  \ElsIf{a supported correction and remaining budget are available}
    \State Apply a bounded update to supported physical controls; retain the update trace
  \Else
    \State Record the rejection or unresolved diagnosis and stop calibration
  \EndIf
\EndWhile
\If{a calibration-accepted candidate was selected}
  \State Freeze its geometry, representation, parameters, and numerical settings
  \State Run the independent probe once; record the physical-verification outcome
\EndIf
\State Associate the separately executed RC outcome by configuration identity
\State \Return Asset record with calibration, independent, and RC evidence
\end{algorithmic}
\end{algorithm}

Ten experts rated 150 core clips and 40 anchors, one rating per clip.
Among 38 clips exceeding a residual-speed ratio of 0.05, 37 were judged
settled. These observations motivate retaining settling as a diagnostic
and using representation-dependent deformation and recovery bands.
The ratings describe perceived plausibility; the study's protocol and
per-dimension counts appear in Appendix~\ref{app:experts}.

\section{Experiments}
\label{sec:experiments}
We evaluate retained coverage, configuration-dependent response, the additional
information from independent testing, and task versus integrity outcomes.
Success rates use all scheduled runs; mean errors use recorded measurements.
Paired statistics resample asset identities across methods and budgets
(Appendix~\ref{app:statistics}).

Table~\ref{tab:evaluation-units} lists the populations behind the results.
The full-pool census and the smaller controlled studies answer different
questions: coverage concerns generated assets, parameter comparisons concern
repeated asset/target evaluations, and grasping concerns condition variants
of a small set of base geometries.

\begin{table}[htbp]
\caption{\label{tab:evaluation-units}\textbf{Evaluation populations and counting units.}
Counts describe reported outcomes. Each full-pool check uses the same assets;
the grasping reference selects one outcome from five closure trials per instance.}
\centering\small
\begin{tabularx}{\linewidth}{@{}lXr@{}}
\toprule
Study & Distinct units and design & Outcomes \\
\midrule
Full-pool verification & 100,000 assets, 485 categories & 100,000 per check \\
Supplementary stability & 2,000 bags, stability drop & 2,000 \\
Garment configurations & 4 garments $\times$ 3 settings $\times$ 2 probes & 24 \\
Response matching & 30 asset/target pairs $\times$ 7 methods $\times$ 4 budgets & 840 \\
Expanded verification & 300 bags $\times$ 7 methods $\times$ 4 budgets & 8,400 \\
Grasping & 1,000 instances from 7 base geometries; three single-trial controllers and a five-trial reference & 4,000 \\
Expert judgments & 150 core clips + 40 anchors; one rating per clip & 190 ratings \\
\bottomrule
\end{tabularx}
\end{table}

\subsection{Coverage survives verification, with uneven retention}
\label{sec:coverage-results}
Physical verification accepts 83.62\% of the pool, RC accepts 99.59\%,
and their intersection retains 83.37\% (Table~\ref{tab:dataset}). All 485
categories remain represented (Figure~\ref{fig:coverage}). The 16,221 assets
that pass RC but miss physical-verification criteria show the additional
requirements of calibration and independent behavioral testing. Family retention ranges from 60.92\% for soft
solids to 98.46\% for cloth sheets.

Soft solids account for 13,493 of 16,317 model-limit rejections (82.69\%).
Low-retention categories include gummy candy (3/206), nama chocolate (9/206),
and toffee (10/207). These counts identify where the present representations
leave sparse coverage. A targeted inspection of 200 rejected objects and
20 controls diagnoses representation failures; a separate 2,000-bag
stability-drop check succeeds in 1,998 cases. Selection criteria and family
outcomes are retained in Appendix~\ref{app:assets}.

\subsection{Garment configurations produce distinct responses}
\label{sec:structure-results}
\begin{figure}[htbp]
\paperfigure{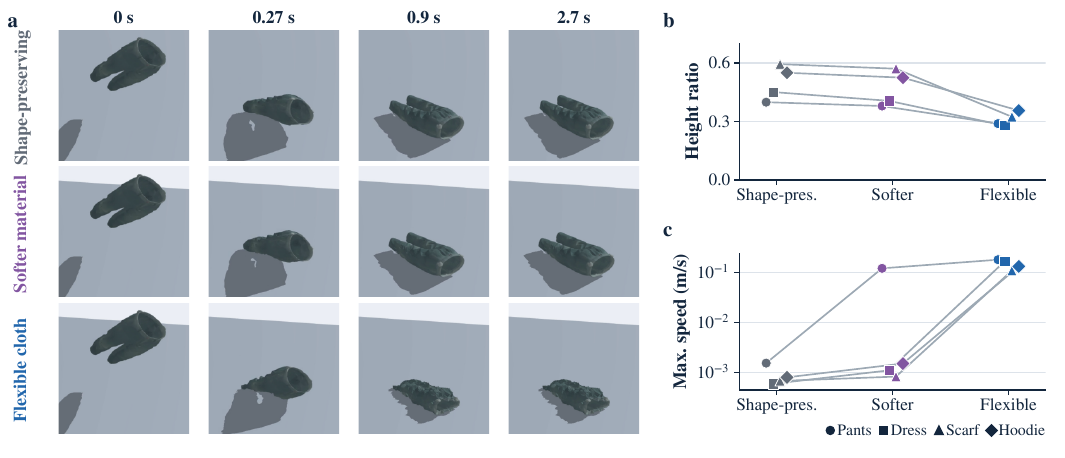}{2in}{Garment configuration responses}
\caption{\label{fig:structure}\textbf{Deformation and residual motion under drop.}
(a) Matched physical times for one garment. (b,c) All four paired height and
speed measurements; speed uses a log scale. Configurations match
Table~\ref{tab:structure-summary}; Appendix~\ref{app:structure} details the settings.}
\end{figure}

Four garments are tested under drop and compression with three configurations:
\emph{shape-preserving}, \emph{softer material}, and \emph{flexible cloth}.
The 24 executions share source geometry and probe timing within each
garment. Configurations jointly change material compliance and structural
support. Table~\ref{tab:structure-summary} gives both actions; Figure~\ref{fig:structure}
shows the paired drop responses.

\begin{table}[htbp]
\caption{\label{tab:structure-summary}\textbf{Garment configuration comparison.}
Each cell averages four garments. Height is final/initial vertical extent;
speed is the final maximum particle speed, averaged across runs.}
\centering\small\setlength{\tabcolsep}{4pt}
\begin{tabular}{lrrrr}
\toprule
& \multicolumn{2}{c}{Drop} & \multicolumn{2}{c}{Compression} \\ Configuration & Height ratio & Speed (m/s) & Height ratio & Speed (m/s) \\
\midrule
Shape-preserving & 0.499 & 0.00091 & 0.876 & 0.15938 \\
Softer material & 0.470 & 0.03125 & 0.707 & 0.00571 \\
Flexible cloth & 0.312 & 0.14820 & 0.303 & 0.15665 \\
\bottomrule
\end{tabular}

\end{table}

The flexible configuration reduces mean drop height ratio to 0.312, compared
with 0.499 for shape preservation and 0.470 for softer material. Under
compression, the corresponding means are 0.303, 0.876, and 0.707. Its drop residual speed also rises
from 0.00091 to 0.148\,m/s. The combined settings permit greater collapse
and longer-lasting residual motion. Individual values and exact settings appear in
Appendix~\ref{app:structure}.

The comparison measures the joint effects of material compliance, structural
support, and scene settings on four garments. Height and residual speed
characterize collapse and motion under the recorded probes.

\subsection{Independent testing reveals calibration-accepted mismatches}
\label{sec:parameter-results}
\paragraph{Evaluation protocol.}
The response-matching experiment uses 30 asset/target pairs, seven methods, and four
allowed budgets ($1,2,4,8$), giving 840 runs. The expanded evaluation uses
300 bags with the same grid, giving 8,400 runs. Calibration compares the
candidate's behavioral summary with the reference response using normalized,
clipped differences; an independent stimulus evaluates the frozen candidate.
The response-matching thresholds are 0.05 for calibration and 0.08 for the
independent test (Appendix~\ref{app:parameters}). The reference is a simulated
response generated with evaluator-side target parameters on the same geometry.
The error therefore measures agreement between simulated behavioral summaries.
Production uses the behavioral bands in Appendix~\ref{app:criteria}.

For a candidate response $c$ and reference response $r$, the adapter compares
the available fields in
\begin{equation}
K\subseteq\{\text{finite fraction},\text{contact fraction},\text{settled},
\text{deformation ratio},\text{recovery ratio}\}.
\end{equation}
A Boolean mismatch contributes zero or one. A numerical field contributes a
clipped relative difference. The recorded error is
\begin{equation}
L(c,r)=\frac{1}{|K|}\sum_{k\in K}d_k,\quad
 d_k=\begin{cases}
 \mathbf{1}[c_k\ne r_k],&\text{Boolean field},\\
 \min\!\left(1,\dfrac{|c_k-r_k|}{\max(|r_k|,10^{-6})}\right),&\text{numerical field}.
 \end{cases}
\label{eq:response-error}
\end{equation}
Fields unavailable in either report are omitted. The adapter returns one if
no fields can be compared. An infrastructure failure with no report remains
missing in the exported error table. Success uses all scheduled runs as its denominator; mean error uses
recorded values and is accompanied by their count.

\begin{table}[htbp]
\caption{\label{tab:verification}\textbf{Calibration and independent-test outcomes.}
``Cal. only'' passes calibration but fails the independent test; ``Indep.
only'' has the reverse outcome. Counts are candidate evaluations across
methods and budgets, with missing measurements kept separate.}
\centering\small\setlength{\tabcolsep}{4pt}
\begin{tabular}{lrrrrrr}
\toprule
Population & Both pass & Cal. only & Indep. only & Neither & Missing & Total \\
\midrule
30 targets & 682 & 49 & 3 & 83 & 23 & 840 \\
300 bags & 8,075 & 325 & 0 & 0 & 0 & 8,400 \\
\bottomrule
\end{tabular}

\end{table}

\paragraph{Additional screening.}
All 8,400 bag evaluations satisfy calibration, but 325 (3.87\%) fail the
independent stimulus (Table~\ref{tab:verification}). The distinct interaction
thus exposes response mismatches after
calibration acceptance. In the 30-target experiment, 49 of 731 observed
calibration-passing evaluations fail independently; 23 runs lack measurements.
Across repeated evaluations, the 325 rejections involve 26 of the 300 bags.
The cross-tabulation measures additional screening of selected candidates.

\paragraph{Candidate executions.}
All expanded runs evaluate the shared initial candidate and pass calibration,
so this experiment measures independent screening. The 30-target study
records 772 one-candidate, 15 two-candidate, and 30 three-candidate runs;
23 runs have missing counts following infrastructure failures. Detailed
local-control and budget summaries appear in Appendix~\ref{app:local-controls}.

\begin{figure}[htbp]
\paperfigure{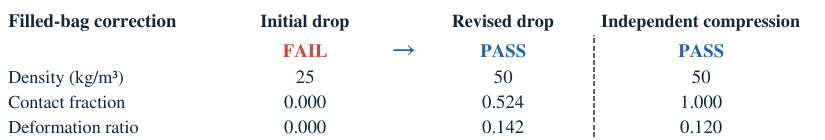}{1in}{A recorded parameter correction}
\caption{\label{fig:feedback}\textbf{A recorded correction and its independent test.}
Observed drop diagnostics lead to a density update. The revised candidate
is frozen before compression. Values are recorded response summaries.}
\end{figure}

\paragraph{Observed feedback corrections.}
Production records contain 83,292 first-candidate acceptances and 326
acceptances after correction, including 209 cloth sheets. A filled-bag case
changes density from 25 to 50\,kg/m$^3$ following a no-contact diagnosis.
The revised drop passes, followed by independent compression with deformation
ratio 0.120 and recovery ratio 1.038 (Figure~\ref{fig:feedback}).
This trace links a contact diagnosis to an executed parameter correction
and subsequent verification.

\subsection{Task completion and a deformation-gradient-based integrity criterion}
\label{sec:application}
A complementary grasp--lift--hold evaluation contains 1,000 qualified
instances drawn from seven base geometries. Fixed command, slip feedback,
a local force prior, and a privileged reference each evaluate the same
660 calibration and 340 held-out instances at seed zero. Base geometries
are shared across the split, which reserves instances and conditions. The
reference selects one outcome from five closure trials using privileged
damage readouts. The protocol comprises 3,000 single-controller trials and
5,000 reference trials, yielding 4,000 reported controller--instance outcomes.

The proxy score $D\in[0,1]$ maps contact-time deformation-gradient statistics
to configured onset/failure bands. Integrity is $1-D$, the damage flag is
$D\geq0.25$, and intact success requires task completion with $D<0.25$.
The score is computed after simulation from $\|F-I\|_F$, which responds
to rotation as well as deformation. Appendix~\ref{app:application} gives
the formula, sampling interval, fallback rules, and thresholds.

\begin{table}[htbp]
\caption{\label{tab:application}\textbf{Final grasping outcomes by instance split.}
Task success is shown for calibration and held-out instances. Intact success
and median damage refer to the 340 held-out instances per method. Intact
success requires task success and damage below 0.25. The last row is a
privileged reference with five closure trials.}
\centering\footnotesize\setlength{\tabcolsep}{3.5pt}
\begin{tabular}{lrrrr}
\toprule
Controller & Cal. task & Held-out task & Intact & Damage \\
\midrule
Fixed command & 595/660 (90.2\%) & 245/340 (72.1\%) & 0/340 & 0.654 \\
Slip feedback & 634/660 (96.1\%) & 245/340 (72.1\%) & 0/340 & 0.784 \\
Force prior (local) & 600/660 (90.9\%) & 200/340 (58.8\%) & 0/340 & 0.478 \\
\midrule
Privileged reference & 360/660 (54.5\%) & 200/340 (58.8\%) & 0/340 & 0.493 \\
\bottomrule
\end{tabular}

\end{table}

\begin{figure}[htbp]
\paperfigure{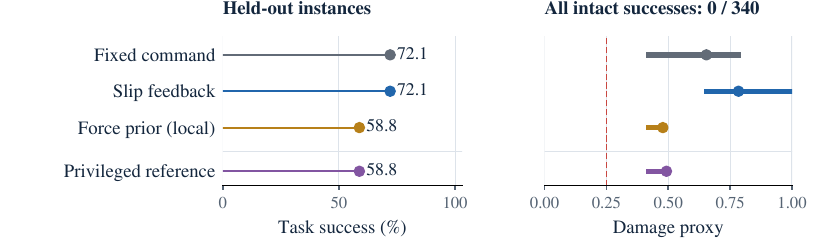}{1.4in}{Task success and integrity}
\caption{\label{fig:application}\textbf{Task success and damage on held-out instances.}
Each controller evaluates 340 instances. Points give task success and median
damage; horizontal segments show damage interquartile ranges. All instances
exceed the damage threshold, so intact success is zero.}
\end{figure}

Across 4,000 recorded outcomes, 3,079 tasks succeed, but all 4,000 trigger the
damage flag and none satisfies intact success. Fixed command and slip
feedback each complete 245/340 held-out tasks, despite different pooled
success counts. Task completion and the specified integrity criterion thus
give different assessments of the same outcomes
(Table~\ref{tab:application}, Figure~\ref{fig:application}).
The minimum observed score is 0.2653, so the chosen 0.25 threshold flags every
outcome. Continuous score distributions describe variation in the proxy
response. Controllers act through commanded closure; their nominal force
settings are mapped to closure depth (Appendix~\ref{app:application}).

\begin{figure}[htbp]
\begingroup
\renewcommand{\paperfigure}[3]{\centering\includegraphics[width=0.90\linewidth]{fig/#1.pdf}}
\paperfigure{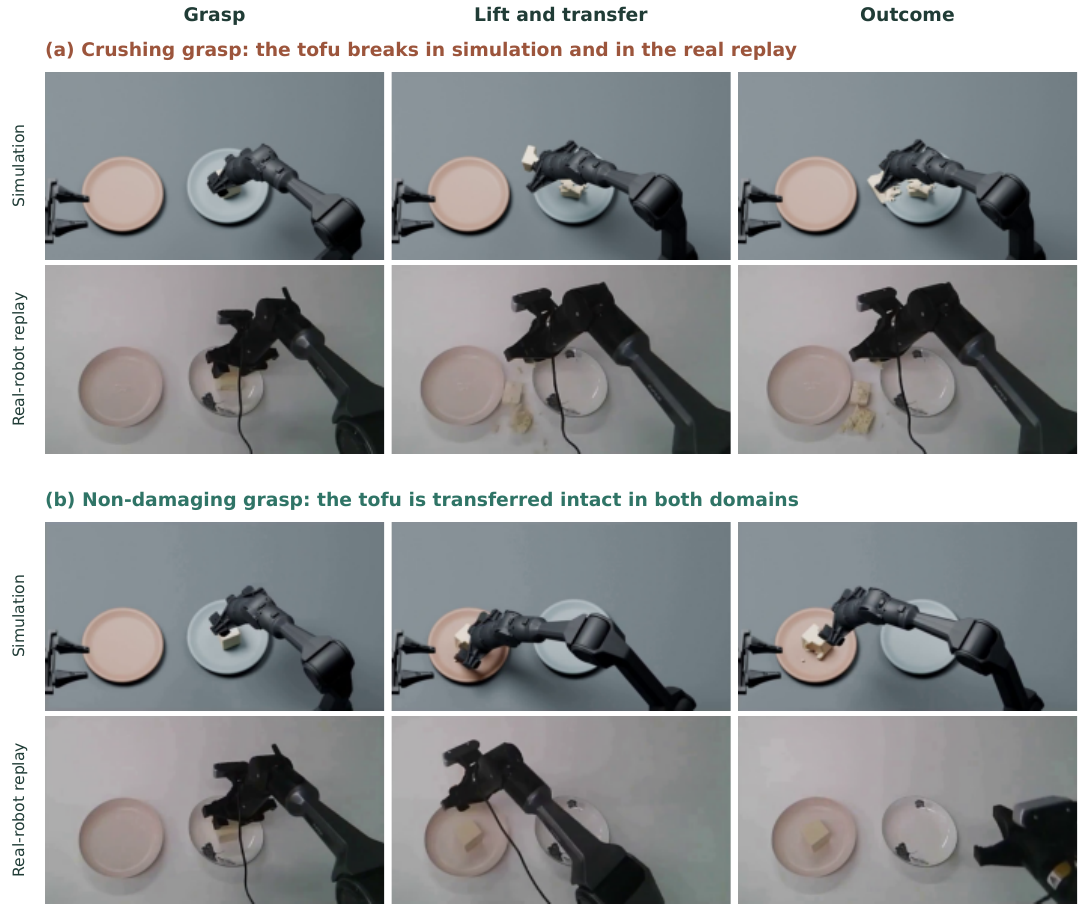}{1in}{Sim-to-real replay of tofu grasping}
\caption{\label{fig:robot-cases}\textbf{Sim-to-real replay of tofu grasping.}
(a) A crushing grasp breaks the tofu in MPM simulation, and replaying the same
gripper closure on the physical Piper arm breaks the real tofu as well.
(b) A non-damaging grasp transfers the tofu intact in both domains.
In each block, simulation frames sit above the real-robot replay at grasp,
lift-and-transfer, and outcome stages; receiving plates appear on the left in
both domains. The matching outcomes show that simulated fragile-object behavior
transfers to the physical robot.}
\endgroup
\end{figure}

\paragraph{Sim-to-real replay of fragile-object manipulation.}
We replay the simulated gripper closure commands on a physical Piper arm to
test whether simulated fragile-object outcomes carry over to the real robot
(Figure~\ref{fig:robot-cases}). The simulation predicts a clear separation
between two grasps of the same tofu block: a crushing grasp breaks the tofu
during lifting, while a non-damaging grasp delivers it to the receiving plate
intact. The real robot reproduces both outcomes. Under the crushing grasp the
real tofu fractures into fragments that fall onto the plate, and under the
non-damaging grasp it is placed without visible breakage. A grasp that crushes
the object in simulation thus also crushes the real object.

The consistent sim-to-real replay indicates that the material response of our
simulated assets is physically accurate: the effective parameters and
representation recorded for the asset reproduce the boundary between damaging
and non-damaging closure observed on the physical robot. This makes the
simulated failure modes informative for real manipulation, complementing the
task and integrity measures above. Appendix~\ref{app:robot-visualization}
describes the replay setup and simulation settings.

\section{Discussion and Limitations}
\label{sec:discussion}

\subsection{Coverage and configuration-specific usability}
RFW-S connects semantic coverage to an executable configuration and behavioral
record for each evaluated asset. Retaining all 485 categories preserves
semantic breadth, while unequal retained counts identify categories with
sparse usable coverage. Targeted representation inspections diagnose selected
failures. Audits of semantic correctness, geometric diversity, and near-duplicate
rates would characterize additional aspects of dataset quality.

Verification results apply to the recorded solver, representation, parameter
configuration, and probes. Effective simulation parameters control those
executions. Paired physical-object measurements would provide an
external reference for response fidelity and inferred scale and material values.

\subsection{Feedback and behavioral criteria}
Independent testing exposes mismatches after calibration acceptance;
production traces document 326 accepted corrections. A controlled study of
correction efficacy would pair update policies on the same initial failures,
retain trace prefixes across budgets, and compare independently tested outcomes
and evaluation cost. The present expanded study evaluates shared initial
candidates across separately executed runs.

Expert judgments inform representation-dependent behavioral bands. The study
uses one rating per clip and summarizes rating distributions and diagnostic
examples. Multiple ratings and a separate validation set would quantify the
reliability of judgments and the agreement between criteria and perceived
plausibility. Historical settling labels are identified separately in
Appendix~\ref{app:probe-cases}.

\subsection{Task completion and the integrity readout}
The grasping application reports task completion together with a specified
integrity criterion. Every reported outcome exceeds the proxy threshold,
limiting comparisons based on binary intact success. The deformation-gradient
readout also responds to rigid rotation. A rotation-invariant strain measure
and calibration against physical damage observations would strengthen its
physical interpretation. Such a change would require recomputing the scores
and outcomes. The sim-to-real replay complements this proxy with a physical
observation: the gripper closure that crushes tofu in simulation crushes the
real tofu, and the non-damaging closure transfers it intact in both domains.
This agreement supports the physical realism of the simulated assets for
fragile manipulation. Paired force and deformation measurements on the
physical robot would further quantify response fidelity across materials.

The instance split shares seven base geometries, and the reference selects
among five trials using privileged readouts. Broader geometry coverage,
additional seeds, and physical response measurements would extend the application
study. Reusable configurations and verification records also support future
interaction and policy-learning experiments under additional stimuli and
solver settings.

\section{Conclusion}
\label{sec:conclusion}
RFW-S connects 100,000 generated assets in 485 everyday deformable categories
to structure-compatible representations, effective parameters, and recorded
behavioral evidence. Physical verification and the separately scheduled RC check
jointly retain 83,365 assets and all categories. Independent testing exposes
calibration-accepted response mismatches, while configuration comparisons
and a grasping application distinguish deformation, task completion, and
a specified integrity criterion. The resulting resource supports reproducible
deformable-object studies with explicit configuration and evaluation scope.
Broader stimuli and paired physical measurements can extend that evidence
toward new interactions and real-object fidelity.

\label{main-text-end}
\FloatBarrier
\bibliographystyle{unsrtnat}
\bibliography{references}
\clearpage
\appendix
\section{Asset Organization and Coverage}
\label{app:assets}
\begin{figure}[htbp]
\conceptfigure{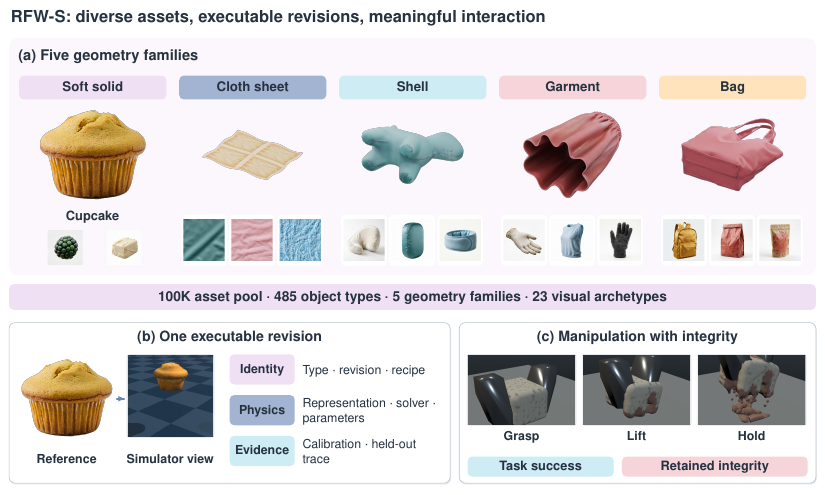}
\caption{\label{fig:overview}\textbf{Asset diversity, executable configurations, and interaction.}
The five structural families connect semantic identity to geometry, physical
parameters, and verification evidence. Reference appearances illustrate coverage, and the simulation schematics
illustrate interaction and integrity criteria.}
\end{figure}

The dataset distinguishes a semantic category, a generated asset, a physical
configuration, and an experimental episode. The census counts 100,000
completed geometric assets; parameter variants and episodes share their
underlying asset identities.
The intersection in Equation~\ref{eq:intersection} is obtained by joining asset,
physical, and RC records and checking their configuration identities.
The category distribution is supplied as a machine-readable table.
Figure~\ref{fig:overview} illustrates the asset organization. The recorded
examples in Figure~\ref{fig:asset-examples} accompany the dataset description.

\begin{table}[htbp]
\caption{Information associated with an evaluated asset configuration.}
\centering\small
\begin{tabularx}{\linewidth}{lX}
\toprule
Component & Contents and purpose \\
\midrule
Semantic identity & Canonical category, structural family, reference appearance \\
Geometry & Source mesh, visual appearance, simulation mesh, repair description \\
Scale & Intended size, solver normalization, coordinate transforms \\
Physical configuration & Representation, solver, proposed and effective parameters \\
Calibration & Probe configuration, measured response, diagnostic, candidate trace \\
Independent verification & Selected configuration, distinct probe, outcome and measurements \\
RC evidence & Execution and behavioral check with matching configuration identity \\
Provenance & File fingerprints linking assets, configurations, and analysis inputs \\
\bottomrule
\end{tabularx}
\end{table}

\begin{table}[htbp]
\caption{Construction outcomes across the complete asset pool. First-candidate
acceptance and acceptance after correction are distinct events.}
\centering\small\begin{tabular}{lr}
\toprule
Construction outcome & Assets \\
\midrule
Accepted at first candidate & 83,292 \\
Accepted after correction & 326 \\
Model-limit rejection & 16,317 \\
Numerical rejection & 2 \\
Unresolved judgment & 63 \\
\bottomrule
\end{tabular}

\end{table}

\begin{table}[htbp]
\caption{\label{tab:family-states}Construction outcomes by structural family.
Corrected acceptances are events with a recorded initial failure and update;
model-limit rejection is the recorded diagnosis.}
\centering\small\setlength{\tabcolsep}{4pt}
\begin{tabular}{lrrrrr}
\toprule
Family & First pass & Corrected & Model limit & Numerical & Unresolved \\
\midrule
Soft solids & 21,071 & 41 & 13,493 & 0 & 36 \\
Cloth sheets & 18,099 & 209 & 36 & 0 & 3 \\
Shells & 11,349 & 66 & 1,553 & 1 & 21 \\
Garments & 19,055 & 0 & 330 & 1 & 0 \\
Bags & 13,718 & 10 & 905 & 0 & 3 \\
\bottomrule
\end{tabular}

\end{table}

\begin{figure}[htbp]
\paperfigure{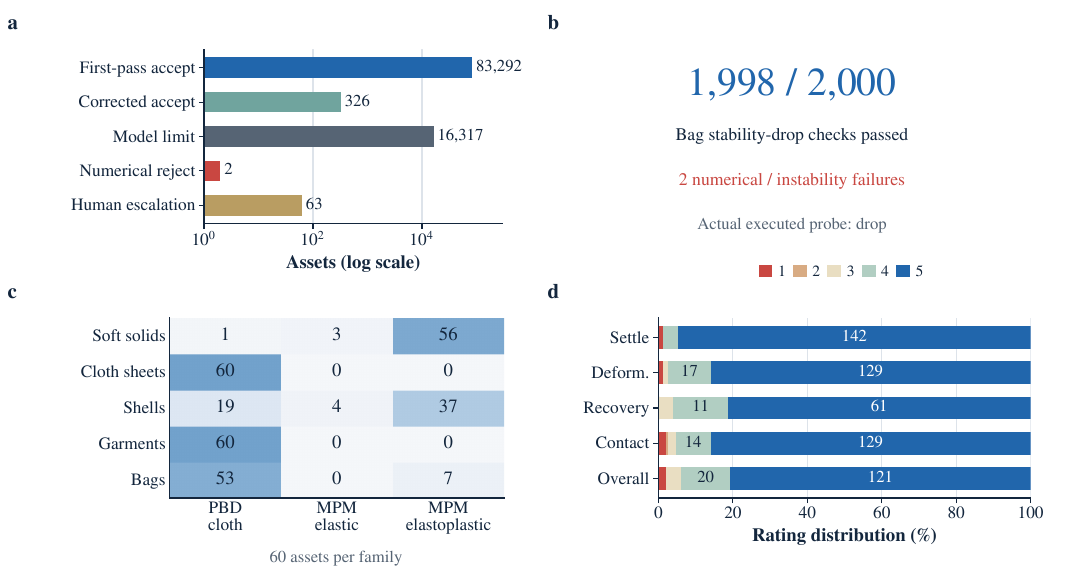}{3in}{Production, solver routing, and expert ratings}
\caption{\label{fig:aux-checks}\textbf{Supplementary verification evidence.}
(a) Production outcomes over 100,000 assets. (b) The 2,000 supplementary bag
runs execute stability-drop checks, with 1,998 passes. (c) Executed solver
modes for 300 assets, 60 per structural family. (d) Ratings of 150 core clips,
with one expert rating per clip; recovery has 75 applicable ratings. Correction counts describe observed construction events.}
\end{figure}

The 300-asset routing analysis connects semantic identity to the solver mode
reported by the executed calibration probe. It contains 193 PBD-cloth, seven
elastic-MPM, and 100 elastoplastic-MPM executions. Table~\ref{tab:routing}
reports their physical verification status. Each solver group contains the objects and representations assigned to it
by the construction procedure. Figure~\ref{fig:aux-checks} summarizes these checks and the expert ratings. The initially proposed route and the executed
mode are kept separately, since structural adaptation can change the route.

\begin{table}[htbp]
\caption{\label{tab:routing}Actual executed solver in the 300-asset routing sample.}
\centering\small\begin{tabular}{lrr}
\toprule
Executed solver & Assets & Physically verified \\
\midrule
PBD cloth & 193 & 189 \\
Elastic MPM & 7 & 6 \\
Elastoplastic MPM & 100 & 74 \\
\bottomrule
\end{tabular}

\end{table}

\section{Probe Metrics and Acceptance Criteria}
\label{app:criteria}
The production record associates each criterion with the executed solver,
probe configuration, and physical configuration identity.
Table~\ref{tab:behavior-criteria} gives the production bands and the RC gate.
Let $\ell$ be the normalized object length, $s_p$ the MPM particle size,
$\Delta x$ the grid spacing, and $h_0$ the initial vertical extent. Contact
uses a distance tolerance $\epsilon_c=\max(3.5s_p,1.25\Delta x)$ for MPM and
$\max(0.02\ell,0.5\Delta x)$ for PBD. Drop contact is detected when a particle
reaches this floor tolerance; compression contact is detected within the
same tolerance of a plate's inner face. The penetration margin is
$\max(s_p,0.02h_0)$ for MPM and $\max(0.02\ell,0.02h_0)$ for PBD.

\begin{table}[htbp]
\caption{\label{tab:behavior-criteria}\textbf{Production behavioral criteria.}
MPM/PBD bands follow the executed representation. Calibration uses drop;
independent verification uses compression--release.}
\centering\normalsize\setlength{\tabcolsep}{4pt}
\begin{tabularx}{\linewidth}{@{}>{\raggedright\arraybackslash}p{0.18\linewidth}>{\raggedright\arraybackslash}X>{\raggedright\arraybackslash}p{0.29\linewidth}@{}}
\toprule
Quantity & Definition & Acceptance \\
\midrule
Finite fraction & Fraction of recorded states with finite positions and velocities & $1.0$ \\
Contact fraction & Fraction of recorded observations with detected contact & $>0$ \\
Drop deformation & $1-h_{\min,\mathrm{contact}}/h_0$ & MPM $[0.005,0.98]$; PBD $[0,0.98]$ \\
Compression deformation & $1-e_{\min}/e_0$, using compression-axis extent & MPM $[0.066,0.95]$ at commanded ratio $0.22$; PBD $[0,0.95]$ \\
Recovery & Final tail mean vertical extent / pre-contact mean vertical extent & MPM $[0.45,1.16]$; PBD $[0,1.16]$ \\
Penetration & Any recorded particle below the negative penetration margin & Absent \\
Settling & Tail mean centroid speed / peak centroid speed & Diagnostic; historical threshold $0.05$ \\
RC outcome & Completed drop execution, finite fraction $1.0$, and drop behavior acceptance & All conditions satisfied \\
\bottomrule
\end{tabularx}
\end{table}

The compression extent uses the 1st--99th particle-coordinate percentiles;
$e_{\min}$ is the minimum during compression and holding. Final recovery uses
the last third of the post-release observations. For a flat proxy whose
vertical reference is below the recorded minimum reference size, recovery
uses final/initial compression-axis extent; the report records this switch.
Settling uses the last 20\% of speed observations, with at least three samples.

The table describes the archived 100,000-asset production reports. A later
revision lowered the MPM compression-deformation bound to $0.02$ for the
complementary grasp-asset qualification after video inspection. Local
response-matching evaluations instead accept $L\leq0.05$ at calibration and
$L\leq0.08$ independently (Equation~\ref{eq:response-error}). They run the
best recorded calibration candidate through the independent probe even when
its calibration error exceeds the threshold; this produces the three
independent-only passes in Table~\ref{tab:verification}. Production
Algorithm~\ref{alg:verification} independently tests calibration-accepted
candidates. Historical qualitative labels are described in
Appendix~\ref{app:probe-cases}.

\section{Structure and Effective Solver Controls}
\label{app:structure}
\subsection{Matched garment configurations}
The four garment cases are 000193, 000202, 002117, and 004801 in the asset
index. Each is evaluated under drop and compression. The three configurations
share the source geometry, initial pose, camera, and a 0.003\,s step with ten
substeps. Their material compliance, rest-curvature support, and floor extents differ.
The three configurations are shape-preserving, softer material, and flexible
cloth; their exact settings accompany the figure-source manifest.

\begin{figure}[htbp]
\paperfigure{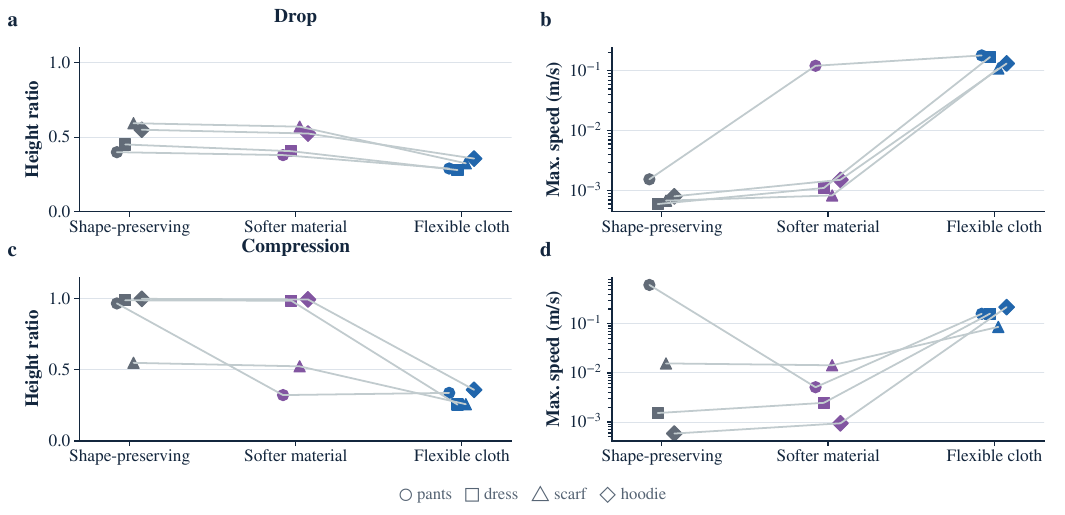}{3in}{All executed garment comparisons}
\caption{\label{fig:structure-all}\textbf{All 24 executed garment comparisons.}
Four garments are evaluated under three configurations and two actions.
Lines pair the same garment across configurations. Height ratio and final
maximum particle speed are separate response measures. The configurations
jointly change compliance, bending iterations, and rest-curvature support.}
\end{figure}

The height ratio is final vertical extent divided by initial vertical extent.
Final maximum speed is the largest recorded particle speed at the final
sample, in m/s. The means in Table~\ref{tab:structure-summary} summarize four
objects each; Table~\ref{tab:structure-individual} retains all paired
measurements. Figure~\ref{fig:structure-all} visualizes the paired values. In particular, the mean speed can be dominated by one moving
object, making individual values useful alongside the mean.

\begin{table}[htbp]
\caption{\label{tab:structure-individual}All garment measurements. Each cell gives
height ratio / final maximum speed (m/s).}
\centering\footnotesize\setlength{\tabcolsep}{4pt}
\begin{tabular}{llrrr}
\toprule
Garment & Probe & Shape-preserving & Softer material & Flexible cloth \\
\midrule
000193 & Drop & 0.399 / 0.002 & 0.379 / 0.122 & 0.289 / 0.181 \\
000193 & Compression & 0.966 / 0.620 & 0.323 / 0.005 & 0.338 / 0.160 \\
000202 & Drop & 0.451 / 0.001 & 0.406 / 0.001 & 0.279 / 0.170 \\
000202 & Compression & 0.988 / 0.002 & 0.985 / 0.002 & 0.256 / 0.161 \\
002117 & Drop & 0.594 / 0.001 & 0.571 / 0.001 & 0.324 / 0.110 \\
002117 & Compression & 0.549 / 0.016 & 0.524 / 0.014 & 0.259 / 0.087 \\
004801 & Drop & 0.550 / 0.001 & 0.525 / 0.002 & 0.356 / 0.133 \\
004801 & Compression & 0.999 / 0.001 & 0.994 / 0.001 & 0.360 / 0.219 \\
\bottomrule
\end{tabular}

\end{table}

The broader representation inspection covers 220 targeted objects: 200
previously rejected examples and 20 controls. Source, drop, and compression
views support inspection of each representation. This targeted sample supports representation diagnosis. A separate
geometric inspection identifies eight of 20 garments as suitable for local
surface repair.

\subsection{Local parameter and observation-window studies}
\label{app:diagnostics}
\begin{figure}[htbp]
\paperfigure{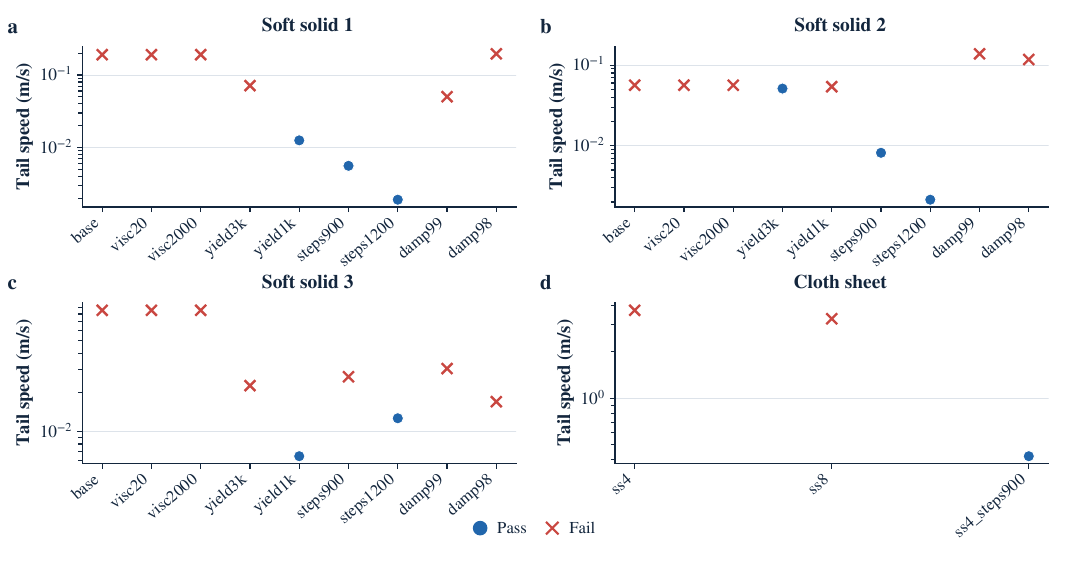}{3in}{Material and observation-window diagnostics}
\caption{\label{fig:local-diagnostics}\textbf{Local solver and observation diagnostics.}
Points show recorded tail mean particle speed. Colors and symbols indicate
the complete behavioral decision, which also uses other measurements.
Variants change material fields, damping, substeps, or observation duration;
each point represents one case at its recorded settings, supplied with the data.}
\end{figure}

A local diagnostic study tests three soft solids and a cloth sheet (Figure~\ref{fig:local-diagnostics}). The
adapter does not pass the proposed plastic-viscosity field to the MPM
material. Altering that field therefore leaves the three solids' measured
tail speeds effectively unchanged: approximately 0.1603, 0.0543, and
0.0853\,m/s across the tested values. A supported yield-stress control changes
the response; for example, the first solid's tail-to-peak speed ratio changes
from 0.147 to 0.008 at the tested lower yield stress. The response is
object-dependent: the second solid remains near the 0.05 diagnostic boundary.
This motivates identifying effective solver controls before using a semantic
parameter proposal as a correction direction.

Observation duration also changes the interpretation of residual motion.
Extending the three solids' drop observations from 420 to 1,200 steps reduces
their respective tail centroid speeds from 0.1603, 0.0543, and 0.0853\,m/s
to 0.0018, 0.0020, and 0.0121\,m/s. The longer observation window resolves
much of the residual motion in these three cases. The cloth checks
likewise show sensitivity to the step/substep configuration. Production
records retain executed step counts, time step, and numerical settings so
that incomplete contact within a short observation window is distinguishable
from an unsupported material response.

\section{Expert Plausibility Judgments}
\label{app:experts}
\begin{figure}[htbp]
\conceptfigure{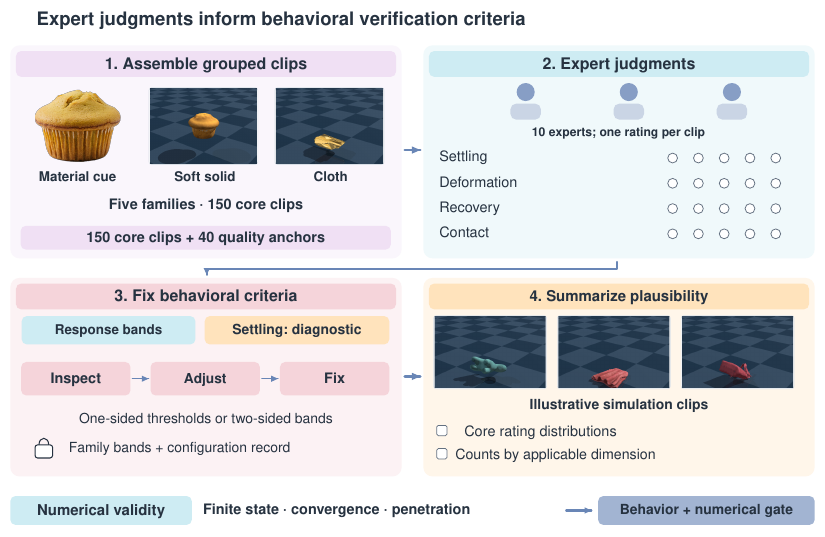}
\caption{\label{fig:human-study}\textbf{Expert plausibility judgments and behavioral criteria.}
Ten experts supplied one rating per clip for 150 core clips and 40 anchors.
Judgments inform representation-dependent response bands; settling remains
a diagnostic. Numerical validity is checked separately; the illustrated
clips explain the assessment workflow.}
\end{figure}

Ten experts each rated 19 clips, giving 190 annotations: 150 core clips and
40 anchors. Each clip received one rating. The 150 core clips cover five
families with 30 clips each, divided equally between drop and compression.
Sampling includes acceptance, boundary, and rejection cases. The core judgments cover settling,
deformation, recovery where applicable, contact, and overall plausibility on
five-point scales. Figure~\ref{fig:human-study} illustrates the study and its use in defining behavioral criteria. Table~\ref{tab:experts} reports the number of ratings of at
least four, with the applicable denominator for each dimension. Recovery has
75 applicable core clips; the other dimensions have 150. The 40 anchors
comprise 20 ordinary examples and 20 deliberately modified low-stiffness
examples for quality checking. Manual review revised ten anchor labels;
all raters were retained. Anchors are excluded from the core proportions
and from the behavioral-band summaries.

\begin{table}[htbp]
\caption{\label{tab:experts}Core expert judgments. Each clip has one expert rating of perceived behavioral plausibility.
Proportions use core clips only, with dimension-specific denominators.}
\centering\small\begin{tabular}{lrrr}
\toprule
Dimension & Rated clips & Score 4 or 5 & Proportion \\
\midrule
Settling & 150 & 148 & 98.7\% \\
Deformation & 150 & 146 & 97.3\% \\
Recovery & 75 & 72 & 96.0\% \\
Contact & 150 & 143 & 95.3\% \\
Overall & 150 & 141 & 94.0\% \\
\bottomrule
\end{tabular}

\end{table}

The judgments identify useful distinctions between numerical diagnostics and
perceived behavior. Among 38 clips exceeding a tail-to-peak speed ratio of
0.05, 37 were judged settled. Recovery could also remain plausible below a
single elastic-style threshold. The core judgments informed wider
drop-deformation and recovery bands and the use of settling as a diagnostic
(Table~\ref{tab:behavior-criteria}). Contact, finiteness, and penetration
criteria retained their numerical definitions.

\section{Parameter Evaluation and Statistics}
\label{app:parameters}
\subsection{Local parameter controls}
\label{app:local-controls}
Table~\ref{tab:parameters} compares fixed profiles, one-shot priors, rule-based
updates, Sobol search, two text-only controls inspired by Vid2Sid and Swim2Real
\citep{qiu2026vid2sid,qiu2026swim2real}, and the full numerical agent, which
uses rules, backtracking, and abstention.
Fixed profiles have lower pooled observed recovery error than the full
agent. Expanded bag pass counts are close across methods: 1,148--1,159 out
of 1,200 each. At budget 8, the paired full-agent minus fixed-profile error
difference is \metric{filter-delta}, with exploratory 95\% interval
\metric{filter-ci}. Every expanded run evaluates one candidate, so these
results characterize candidate outcomes. Budgets are separate executions;
actual counts and paired intervals are provided in Appendix~\ref{app:statistics}.

Methods share identical initial parameter values within each asset/budget
group. All expanded runs stop calibration at that initial candidate, so
method-labelled differences summarize separate executions of shared starting
configurations. In the 30-target records, three runs show a parameter change;
the one run labelled corrected acceptance reports no parameter change.
Candidate counts include repeated evaluations and retries.

The plot labels RuleTune and Seeded Sobol refer to rule-based updates and
Sobol search. Vid2Sid-style and Swim2Real-style are local text-only controls;
the latter adds backtracking. The full numerical agent uses rules,
backtracking, and abstention. These local controls share our parameter
interface and recorded probes. Production additionally performs structural
adaptation and construction decisions.

\begin{table}[htbp]
\caption{\label{tab:parameters}\textbf{Seven-method independent evaluation.}
Recovery uses 30 targets and 120 runs per method; the expanded set uses
300 bags and 1,200 runs per method. Errors are means over recorded values
(lower is better). All expanded errors are observed; recovery valid counts
expose missingness. The two language-model controls are local text-only adaptations.}
\centering\footnotesize\setlength{\tabcolsep}{3.5pt}
\begin{tabular}{lrrrrr}
\toprule
Method & Recovery pass & Error & Valid n & Expanded pass & Error \\
\midrule
Fixed profile & 100/120 & 0.0521 & 115 & 1152/1200 & 0.0127 \\
One-shot prior & 99/120 & 0.0704 & 119 & 1148/1200 & 0.0128 \\
Rule-based update & 98/120 & 0.0626 & 117 & 1151/1200 & 0.0125 \\
Sobol search & 97/120 & 0.0670 & 117 & 1156/1200 & 0.0121 \\
Vid2Sid-style (text) & 98/120 & 0.0622 & 115 & 1151/1200 & 0.0128 \\
Swim2Real-style (text) & 96/120 & 0.0734 & 118 & 1159/1200 & 0.0118 \\
Full numerical agent & 97/120 & 0.0691 & 116 & 1158/1200 & 0.0117 \\
\bottomrule
\end{tabular}

\end{table}

\subsection{Recorded error and acceptance}
The response error $L(c,r)$ is defined in Equation~\ref{eq:response-error}.
Success uses the scheduled-run denominator; mean error uses recorded
measurements, with valid counts reported separately.

\begin{figure}[htbp]
\paperfigure{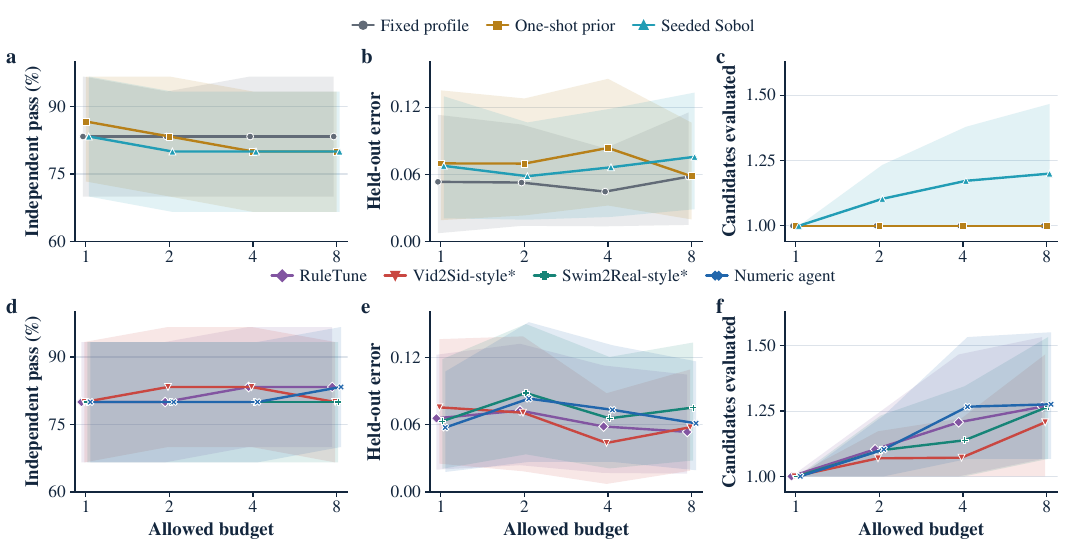}{3in}{Recovery outcomes and candidate budgets}
\caption{\label{fig:recovery-budgets}\textbf{Recovery evaluation across allowed budgets.}
Columns show independent-pass fraction, observed behavior-summary error,
and candidates evaluated; rows separate method groups. Budgets are separate runs.
The 840 runs include 685 successes and 23 infrastructure failures with missing
errors. Failures remain in success denominators. Whiskers are 95\% percentile
intervals from 5,000 asset/target bootstrap resamples with shared draws across
methods and budgets. Asterisks mark local text-only language-model adaptations.}
\end{figure}

\begin{figure}[htbp]
\paperfigure{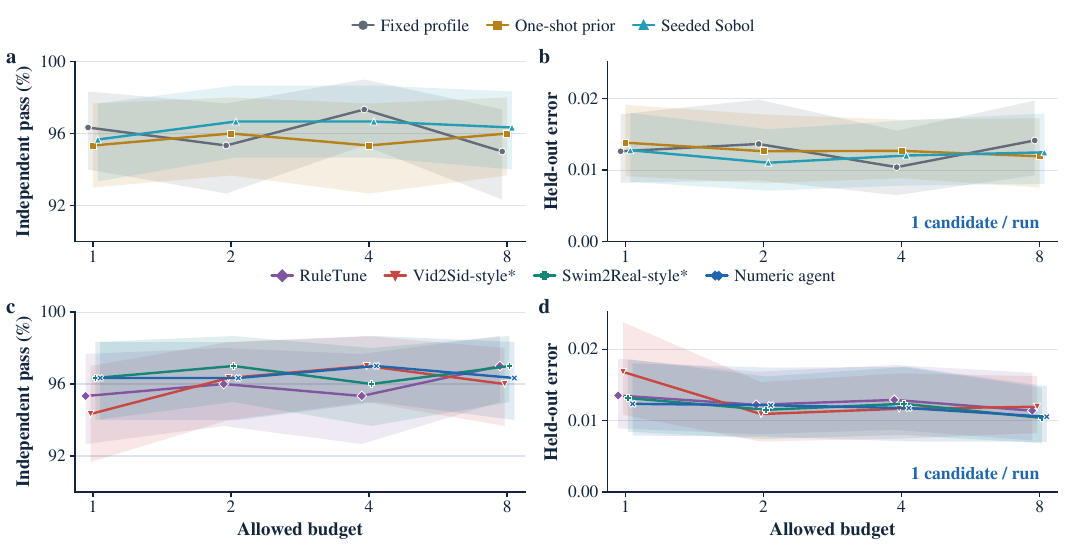}{3in}{Independent bag verification across budgets}
\caption{\label{fig:filter-budgets}\textbf{Expanded bag verification across allowed budgets.}
The 8,400 runs cover 300 bags, seven methods, and four allowed budgets;
8,075 pass independent verification. All errors are recorded and every run
evaluates one candidate. Columns show pass fraction and response error; rows
separate method groups. Bootstrap intervals follow Figure~\ref{fig:recovery-budgets}.
Each point summarizes candidate outcomes at the indicated allowed budget.}
\end{figure}

The recovery evaluation includes 30 assets and 30 target handles. Each
method/budget pair schedules those 30 targets; the four budgets are separate
runs. Of 840 runs, 772 evaluate one candidate, 15 evaluate two, 30 evaluate
three, and 23 lack this measurement following infrastructure failures.
Figures~\ref{fig:recovery-budgets} and~\ref{fig:filter-budgets} show acceptance,
error, and actual candidate counts alongside the allowed budgets.
The expanded bag evaluation schedules 300 assets per method/budget pair, and
all 8,400 runs evaluate one candidate. Its independent check accepts 8,075 candidates and rejects 325.

\begin{table}[htbp]
\caption{\label{tab:recovery-budgets}Recovery error by allowed candidate budget.
Parentheses give the number of recorded errors out of 30 scheduled targets.}
\centering\footnotesize\setlength{\tabcolsep}{4pt}
\begin{tabular}{lrrrr}
\toprule
Method & Budget 1 & Budget 2 & Budget 4 & Budget 8 \\
\midrule
Fixed profile & 0.0533 (28) & 0.0527 (29) & 0.0446 (29) & 0.0579 (29) \\
One-shot prior & 0.0698 (30) & 0.0696 (30) & 0.0836 (30) & 0.0582 (29) \\
Rule-based update & 0.0659 (29) & 0.0727 (29) & 0.0585 (29) & 0.0537 (30) \\
Sobol search & 0.0676 (29) & 0.0583 (29) & 0.0663 (29) & 0.0757 (30) \\
Vid2Sid-style (text) & 0.0756 (29) & 0.0709 (29) & 0.0438 (28) & 0.0578 (29) \\
Swim2Real-style (text) & 0.0633 (29) & 0.0884 (30) & 0.0658 (29) & 0.0755 (30) \\
Full numerical agent & 0.0576 (28) & 0.0834 (29) & 0.0734 (30) & 0.0614 (29) \\
\bottomrule
\end{tabular}

\end{table}
\begin{table}[htbp]
\caption{\label{tab:filter-budgets}Expanded bag verification error by allowed
budget. Every cell contains 300 recorded values; all runs evaluate one candidate.}
\centering\footnotesize\setlength{\tabcolsep}{4pt}
\begin{tabular}{lrrrr}
\toprule
Method & Budget 1 & Budget 2 & Budget 4 & Budget 8 \\
\midrule
Fixed profile & 0.0126 (300) & 0.0136 (300) & 0.0104 (300) & 0.0141 (300) \\
One-shot prior & 0.0138 (300) & 0.0126 (300) & 0.0127 (300) & 0.0119 (300) \\
Rule-based update & 0.0135 (300) & 0.0122 (300) & 0.0129 (300) & 0.0114 (300) \\
Sobol search & 0.0128 (300) & 0.0110 (300) & 0.0120 (300) & 0.0125 (300) \\
Vid2Sid-style (text) & 0.0168 (300) & 0.0109 (300) & 0.0116 (300) & 0.0119 (300) \\
Swim2Real-style (text) & 0.0131 (300) & 0.0115 (300) & 0.0123 (300) & 0.0104 (300) \\
Full numerical agent & 0.0123 (300) & 0.0121 (300) & 0.0117 (300) & 0.0105 (300) \\
\bottomrule
\end{tabular}

\end{table}

\subsection{Calibration and independent-test cross-tabulation}
The calibration response tolerance is 0.05 and the independent response
tolerance is 0.08 in the executed evaluation adapter. A run with missing
calibration or independent error is listed separately. The cross-tabulation
joins both measurements within the same candidate evaluation, and the
independent-pass column follows its recorded final outcome. In the 30-target
experiment, 685 independent passes include three calibration failures;
682 satisfy both response criteria. The expanded experiment has 8,075 joint
passes and 325 calibration-only passes involving 26 distinct bags.

\begin{table}[htbp]
\caption{\label{tab:verification-methods}Cross-tabulation by method.
``Both'' passes both response criteria; the two ``only'' columns pass exactly
one. Missing executions remain outside the four observed outcome cells.}
\centering\footnotesize\setlength{\tabcolsep}{3pt}
\begin{tabular}{llrrrrr}
\toprule
Experiment & Method & Both & Cal. only & Indep. only & Neither & Missing \\
\midrule
Recovery & Fixed profile & 99 & 4 & 1 & 11 & 5 \\
Recovery & One-shot prior & 98 & 5 & 1 & 15 & 1 \\
Recovery & Rule-based update & 97 & 6 & 1 & 13 & 3 \\
Recovery & Sobol search & 97 & 9 & 0 & 11 & 3 \\
Recovery & Vid2Sid-style (text) & 98 & 10 & 0 & 7 & 5 \\
Recovery & Swim2Real-style (text) & 96 & 10 & 0 & 12 & 2 \\
Recovery & Full numerical agent & 97 & 5 & 0 & 14 & 4 \\
Filter & Fixed profile & 1152 & 48 & 0 & 0 & 0 \\
Filter & One-shot prior & 1148 & 52 & 0 & 0 & 0 \\
Filter & Rule-based update & 1151 & 49 & 0 & 0 & 0 \\
Filter & Sobol search & 1156 & 44 & 0 & 0 & 0 \\
Filter & Vid2Sid-style (text) & 1151 & 49 & 0 & 0 & 0 \\
Filter & Swim2Real-style (text) & 1159 & 41 & 0 & 0 & 0 \\
Filter & Full numerical agent & 1158 & 42 & 0 & 0 & 0 \\
\bottomrule
\end{tabular}

\end{table}

\subsection{Paired analysis and missingness}
\label{app:statistics}
The analysis unit is an asset/target, with methods paired within each budget.
For a paired error difference, both methods must have a recorded value.
We draw 5,000 bootstrap resamples of asset IDs with replacement, using seed
20260926 and the same resampling indices across methods and budgets within
each experiment. Each replicate averages the available paired differences;
its 2.5th and 97.5th percentiles define the reported interval. The count of
observed pairs is reported explicitly. This retains the population's missing
pair positions while conditioning the error estimate on observed pairs.

\begin{table}[htbp]
\caption{\label{tab:paired}Exploratory paired differences: full numerical agent
minus fixed profile. Negative values favor the numerical agent on recorded
error. Intervals are unadjusted for multiple comparisons.}
\centering\small\begin{tabular}{lrrrr}
\toprule
Experiment & Budget & Pairs & Difference & 95\% interval \\
\midrule
Recovery & 1 & 27 & -0.0102 & [-0.0355, 0.0113] \\
Recovery & 2 & 28 & 0.0227 & [-0.0028, 0.0593] \\
Recovery & 4 & 29 & 0.0196 & [-0.0015, 0.0544] \\
Recovery & 8 & 28 & 0.0052 & [0.0000, 0.0117] \\
Filter & 1 & 300 & -0.0003 & [-0.0043, 0.0037] \\
Filter & 2 & 300 & -0.0015 & [-0.0048, 0.0018] \\
Filter & 4 & 300 & 0.0013 & [-0.0026, 0.0056] \\
Filter & 8 & 300 & -0.0036 & [-0.0071, -0.0004] \\
\bottomrule
\end{tabular}

\end{table}

The paired estimates summarize differences between separately executed
method-labelled runs under the recorded candidate counts.

\section{Complementary Grasp Evaluation}
\label{app:application}
The construction produces 2,000 candidate instances, of which 1,000 qualify
for evaluation. The instances comprise parameter and condition variants of seven base geometries. Each controller
uses the same 660 calibration and 340 held-out instances at seed zero.
The split partitions instances, including variants of shared base geometries.
Qualification requires a verified physical configuration. Three controllers
execute one trial per instance; the reference protocol executes five and
selects one outcome. Thus the protocol schedules 8,000 trials and reports
4,000 controller--instance outcomes.

\paragraph{Controllers and evaluation access.}
Fixed command follows the prescribed closure. Slip feedback tightens the
closure by 2\% of initial width when vertical sag exceeds 3\,mm, bounded
by a minimum gap of 60\% of its initial value. The local force prior supplies
a nominal force $f$ in newtons, mapped to closure ratio
$\operatorname{clip}_{[0.2,0.6]}(f/60)$ by the interface. These are kinematic
commands; measured contact force is unavailable in the reported records.

The privileged reference executes closure ratios 0.10, 0.20, 0.30, 0.45,
and 0.60. It selects the lowest-score intact task success when available;
otherwise it selects the lowest-score trial among all five executions,
including task failures. The reference row therefore reports damage-based
selection, which can select a task failure despite a successful alternative.

\paragraph{Deformation-gradient-based proxy readout.}
Let $F_i(t)$ denote the deformation gradient of particle $i$, and let
$\mathcal T_c$ contain sampled observation times at which contact is detected.
Every twentieth observation with contact contributes to
\begin{equation}
q=\max_{t\in\mathcal T_c}\operatorname{Q}_{0.95,i}
       \bigl(\|F_i(t)-I_3\|_F\bigr),\qquad
D=\operatorname{clip}_{[0,1]}\!\left(\frac{q-a}{b-a}\right),
\label{eq:damage}
\end{equation}
where $I_3$ is the identity matrix and $(a,b)$ are the onset/failure bands.
Low, medium, and high fragility settings use $(0.35,0.75)$, $(0.25,0.60)$,
and $(0.15,0.45)$, respectively. The implementation falls back to the final particle mean of $\|F_i-I_3\|_F$ if no positive contact-time peak is available, and then to an
aperture/extent statistic if the deformation gradient is unavailable.
The resulting $D$ is a normalized proxy from the simulated response. For a
pure rigid rotation $F_i=R$, $\|R-I_3\|_F$ can be positive even with no
material strain. The readout therefore includes rotation as well as
deformation. It is evaluated after simulation as a response readout; the
simulator evolves the configured material independently of this score.

Integrity is $1-D$. A damage flag is triggered when $D\geq0.25$; intact
success requires task completion and $D<0.25$. Task success requires the
centroid to reach at least 80\% of the commanded 30\,mm lift, to retain at
least 25\,mm elevation during holding, and to remain finite. The proxy is
computed from the executed MPM/PBD response as an evaluation readout.

\begin{table}[htbp]
\caption{\label{tab:application-detail}All application splits.
Task, intact success, and damage flag are counts; damage and integrity are
medians. The flag overlaps task success. ``All'' pools the two instance
splits; each controller has the same underlying instances.}
\centering\footnotesize\setlength{\tabcolsep}{3pt}
\begin{tabular}{llrrrrrr}
\toprule
Split & Controller & N & Task & Intact & Flag & Damage & Integrity \\
\midrule
Calibration & Fixed command & 660 & 595 & 0 & 660 & 0.658 & 0.342 \\
Held-out & Fixed command & 340 & 245 & 0 & 340 & 0.654 & 0.346 \\
All & Fixed command & 1000 & 840 & 0 & 1000 & 0.654 & 0.346 \\
Calibration & Slip feedback & 660 & 634 & 0 & 660 & 0.950 & 0.050 \\
Held-out & Slip feedback & 340 & 245 & 0 & 340 & 0.784 & 0.216 \\
All & Slip feedback & 1000 & 879 & 0 & 1000 & 0.784 & 0.216 \\
Calibration & Force prior (local) & 660 & 600 & 0 & 660 & 0.503 & 0.497 \\
Held-out & Force prior (local) & 340 & 200 & 0 & 340 & 0.478 & 0.522 \\
All & Force prior (local) & 1000 & 800 & 0 & 1000 & 0.489 & 0.511 \\
Calibration & Privileged reference & 660 & 360 & 0 & 660 & 0.501 & 0.499 \\
Held-out & Privileged reference & 340 & 200 & 0 & 340 & 0.493 & 0.507 \\
All & Privileged reference & 1000 & 560 & 0 & 1000 & 0.493 & 0.507 \\
\bottomrule
\end{tabular}

\end{table}

All 4,000 records contain valid damage and integrity measurements and report
convergence. There are 3,079 task successes and 921 task failures; all 921
failures carry the archived damage-failure reason. This logging category
is assigned to a task failure when the damage flag is present. Independently,
the damage flag is present in all 4,000 outcomes, and intact success is zero.
Failure reasons partition task failures, while damage flags apply to all outcomes.
Across all instances, fixed command, slip feedback, force prior, and the
privileged reference have 840, 879, 800, and 560 task successes, respectively.
The held-out results are 245, 245, 200, and 200 out of 340. The split-specific results distinguish the equal held-out performance
of the first two controllers from their different pooled counts.
This complementary collection evaluates the construction interface and
its task and integrity measures.

\subsection{Sim-to-real replay protocol}
\label{app:robot-visualization}
Figure~\ref{fig:robot-cases} pairs two MPM simulations with two physical-robot
recordings of the same tofu grasp--transfer task. The gripper closure commands
of each simulated grasp are replayed on a physical Piper arm. The simulation
uses a dual-arm Piper scene with the right arm active and the left arm
stationary; the figure shows the manipulation area. An oblique orthographic
camera approximates the recording viewpoint, and representative frames are
selected by action stage. Robot frames are cropped and proportionally resized,
without mirroring or synthesized intermediate frames.

Both simulated cases use 6,460 MPM particles and identical material settings:
Young's modulus 28\,kPa, Poisson ratio 0.32, density 950\,kg/m$^3$, yield
stress 5\,kPa, and a damage extension that reduces shear and tensile strength
over an accumulated-plastic-strain interval of 0.15--0.45. The two cases differ
only in the commanded gripper motion. In the crushing case the gripper holds
the block at the pickup plate, closes further, lifts, and translates; local
damage develops before lifting, the block loses support and falls, and the
impact adds deformation. In the non-damaging case the gripper lifts and
transfers the block with moderate closure. The contrast between breakage and
intact transfer therefore follows from the commanded closure rather than from
different material settings, which is the quantity replayed on the robot.

Finger-mesh signed distance fields support particle contact projection and
friction, and the receiving surface is approximated by a plane. Reconstructed
particle surfaces are excluded from the finger interiors for rendering. The
MPM damage model is separate from the post-hoc proxy in
Equation~\ref{eq:damage}; the replay outcomes are reported in
Section~\ref{sec:application}.

\section{Reproducibility and Implementation Details}
\label{app:implementation}
\subsection{Model roles and inference configuration}
\label{app:backbones}
Section~\ref{sec:model-roles} and Table~\ref{tab:model-config} identify the
models, input modalities, and sampling temperatures. Physical-prior records
store the API alias as their revision identifier. Local-control transport
permits runtime overrides, while canonical evaluations retain configuration
information, with per-request model identities absent from the canonical records. The following details
specify the prompt and budget configuration.

\paragraph{Prompts and parameter constraints.}
The physical-prior configuration sets an 8,192-token output limit and
disables the API's thinking option. Its prompt version is
\path{softsim-physical-production-v1-qwen3.8-max}. Reference
images accompany structured semantic and parameter-catalog information.
Outputs are uncertain simulation priors, including parameter values,
confidence, and rationale. Production correction uses a separate JSON-only
prompt requesting one physical-parameter patch and a rationale from the
current prior, candidate, and calibration metrics/diagnosis. The adapter
restricts changes to supported physical parameters and enforces bounds;
solver time step, substeps, and other numerical settings remain fixed.
The source package records the prompt text, selected configuration fields,
and their source fingerprints.

\paragraph{Budgets and local comparison scope.}
The production configuration starts with three candidates and allows
extensions by two, up to eight, while relative calibration improvement meets
the 0.05 threshold. Independent verification follows candidate selection.
The parameter comparisons instead use separately scheduled budgets of
1, 2, 4, and 8, measured in simulator candidate evaluations.
The text-only controls share the configured Qwen transport, whereas the
Numeric agent uses mechanical updates, backtracking with factor 0.5, and
abstention after two consecutive non-improving updates. Candidate execution
counts and parameter-change records appear in Appendix~\ref{app:local-controls}.

\subsection{Records and analysis reproduction}
\paragraph{Inputs and record identity.}
The census joins the asset manifest, physical-configuration index, and full
runtime records. Asset and physical-configuration identities match across
these inputs. The five complete evaluations contain 100,000 runtime records,
2,000 supplementary stability checks, 840 recovery records, 8,400 expanded
verification records, and 4,000 application records. Completion metadata,
task manifests, and canonical records have matching counts and recorded file
fingerprints. Rejected and infrastructure-failure records remain present.

\paragraph{Analysis reproduction.}
The accompanying analysis package supplies machine-readable category and
family counts, construction states, per-run parameter measurements,
application outcomes, and their sources. The manuscript's table generator
produces the result macros and tables from these files. Paired statistics
use the same resampling scheme as the visualization scripts. Figure inputs
record the source asset or episode, simulation configuration, file
fingerprint, and frame time when applicable. The source package preserves
explicit figure identities and includes the final figure files, numerical inputs, and their fingerprints.

\paragraph{Scope of auxiliary evidence.}
The 2,000-record supplementary check evaluates bag stability under drop.
The structure study, expert judgments, local parameter diagnostics, and
application collection each use their specified selection criteria and
denominators, retaining infrastructure failures in the scheduled-run counts.
Software fixtures and execution pilots are tracked separately.

\clearpage
\section{Qualitative Probe Outcomes}
\label{app:probe-cases}
\begin{figure}[H]
\paperfigure{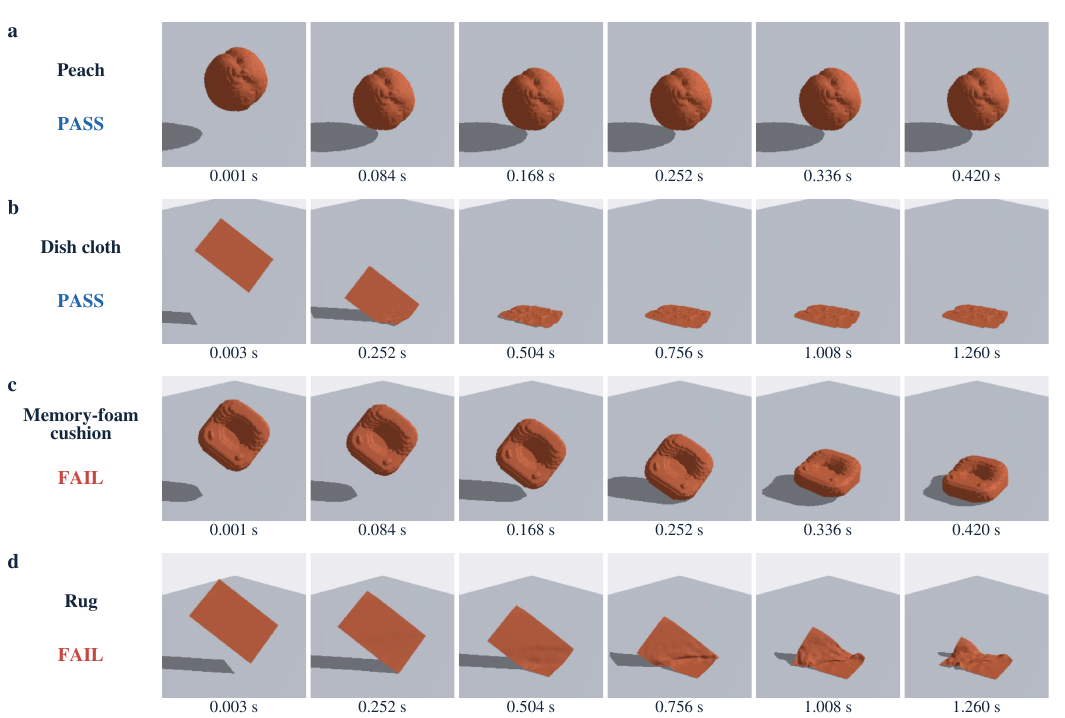}{0.48\textheight}{Drop probe outcomes}
\caption{\label{fig:drop-cases}\textbf{Drop tests: passing and non-settling cases.}
Rows (a, b) show passing peach and dish-cloth runs; rows (c, d) show
memory-foam-cushion and rug runs that fail the recorded settling criterion.
Frames are sampled at steps 1, 84, 168, 252, 336, and 420.
Physical times use the effective solver step: 0.001\,s in (a, c) and
0.003\,s in (b, d). The first frame follows the first simulation step.
The camera and crop remain fixed within each row; object scales differ
across rows. The orange appearance comes from the original simulation videos.}
\end{figure}

\begin{figure}[H]
\paperfigure{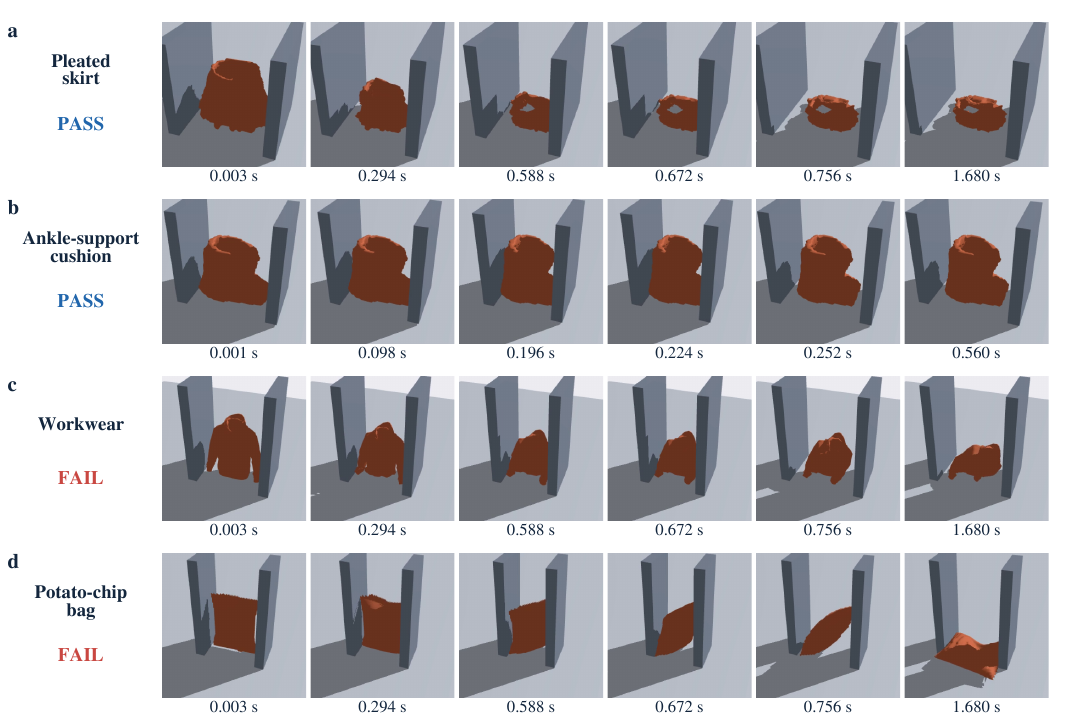}{0.48\textheight}{Compression--release probe outcomes}
\caption{\label{fig:compression-cases}\textbf{Compression--release tests:
passing and non-settling cases.}
Rows (a, b) show passing pleated-skirt and ankle-support-cushion runs;
rows (c, d) show workwear and potato-chip-bag runs that fail the recorded
settling criterion. All four trajectories remain finite with no recorded
ground penetration; FAIL denotes residual motion after release.
Compression, holding, and retraction last 196, 28, and 28 steps, respectively.
Frames show steps 1, 98, 196, 224, 252, and 560, using a 0.001\,s solver
step in (b) and 0.003\,s in (a, c, d). Each row retains its fixed camera
and full original frame; object scales differ across rows.}
\end{figure}

Figures~\ref{fig:drop-cases} and~\ref{fig:compression-cases} show eight completed
drop and compression--release runs across the five structural families.
PASS/FAIL reproduces the automated behavior decision attached to each recorded
video. All four FAIL cases retain motion at the end of the observation window
and fail its settling criterion. The labels refer to the settling-based
assessment used for these recordings; the final acceptance criteria instead
treat settling as a diagnostic (Appendix~\ref{app:experts}). Each row shows
a distinct asset and configuration, selected to illustrate passing or
non-settling behavior.

\subsection{Drop}

\clearpage
\subsection{Compression--release}

\section{Future Work}
\label{app:future}

A natural extension of RFW-S is toward hybrid articulated--deformable
objects, where rigid kinematic structures are coupled with soft components
such as fabric covers, straps, cushions, seals, membranes, or compliant
connectors. Extending the current structure-aware construction and independent
verification procedure to such mixed systems could bridge deformable-object
simulation with articulated reconstruction, part-level kinematics, and
physically grounded interaction
\citep{sapien2020,gapartnet2022,gaussianart2025,
yang2026sketch2arti,huang2026arthoi,cao2026physxomni,chen2026unigarment,lyu2026rfwa}.

At the scene level, RFW-S assets could serve as physically grounded building
blocks for generative simulation. Future systems could automatically compose
editable and executable worlds with scene layouts, lighting, articulated and
deformable objects, and task-relevant interactions, supporting scalable data
generation as well as controlled construction of rare or challenging scenarios
\citep{deitke2022procthor,yang2024holodeck,idea23d2024,wang2024gensim,wang2024robogen,
xu2025cruise,li2026genncap,chen2026lumera,chen2026ovow,xu2025diffusionknowstransparency,shuai2026gennav}.

Beyond explicit simulation, the physical configurations and executed
interaction traces in RFW-S could support 4D and physics-aware video
generation. Coupling simulator states with generative video renderers could
provide controllable dynamics under object motion, camera trajectories,
illumination, and external interactions, while video-to-4D reconstruction
could close the loop by recovering structured, simulation-ready worlds from
real or generated observations
\citep{ren2023dreamgaussian4d,zhang2024physdreamer,liu2026lightx,
chen2026ovow,chen2026dar}.

Finally, RFW-S could support broader embodied benchmarks and policy learning
beyond the grasp--lift--hold protocol studied here, including long-horizon
deformable manipulation, bimanual and high-DoF dexterity, VLA training,
safety-aware interaction, and sim-to-real evaluation. Its physical
verification and integrity records could enable evaluation criteria that
jointly consider task completion, object preservation, and physical or
semantic safety
\citep{lin2020softgym,lu2024garmentlab,gu2023maniskill2,
liu2023libero,kim2024openvla,zhang2026dexora,cui2026liberosafety,
ma2024eureka,ma2024dreureka,rise2026,unidex2026}.
\end{document}